\RequirePackage{fix-cm}
\documentclass[smallextended]{svjour3}       
\usepackage{amsthm}
\usepackage{amssymb}
\usepackage{graphicx}
\usepackage{cite}
\usepackage{hyperref}
\newtheorem{Proposition}{Proposition}
\newtheorem{Property}{Property}
\newtheorem{Definition}{Definition}

\renewenvironment{proof}{\noindent{ \emph{Proof.} }}{\qedsymbol}
\usepackage{algorithm}
\usepackage{algorithmic}
\usepackage{bbding}
\usepackage{pifont}
\usepackage{wasysym}
\usepackage[dvipsnames]{xcolor}
\usepackage{array}
\usepackage{multirow}
\usepackage{float}
\newcolumntype{M}[1]{>{\centering\arraybackslash}m{#1}}

\journalname{Applied Intelligence}
\begin{document}

\title{Data Obsolescence Detection in the Light of Newly Acquired Valid Observations
}


\author{Salma Chaieb        \and
        Brahim Hnich \and
        Ali Ben Mrad 
}


\institute{S. Chaieb \at
              University of Sousse, ISITCom, 4011, Sousse, Tunisia \\
              University of Sfax, CES Lab, 3038, Sfax, Tunisia \\
              Tel.: +216 53 950 523\\
              \email{salma.chaieb2@yahoo.com}  
         \and
           B. Hnich \at
           University of Monastir, FSM, 5000, Monastir, Tunisia \\
           University of Sfax, CES Lab, 3038, Sfax, Tunisia 
         \and
           A. Ben Mrad \at
           University of Sfax, ISAAS, 1013, Sfax, Tunisia \\
           University of Sfax, CES Lab, 3038, Sfax, Tunisia
}

\date{Received: date / Accepted: date}

\maketitle

\begin{abstract}
The information describing the conditions of a system or a person is constantly evolving and may become obsolete and contradict other information. A database, therefore, must be consistently updated upon the acquisition of new valid observations that contradict obsolete ones contained in the database. In this paper, we propose a novel causation-based system for dealing with the information obsolescence problem when a causal Bayesian network is our representation model. Our approach is based on studying causal dependencies between the network variables to detect, in real-time, contradictions between the observations on a single subject and then identify the obsolete ones. We propose a new approximate concept, $\epsilon$-Contradiction, which represents the confidence level of having a contradiction between some observations relating to a specific subject. Once identified, obsolete observations are given in an original way, in the form of an explanation AND-OR Tree. Our approach can be applied in various domains where the main issue is to detect and explain  personalized situations such that the reasons and circumstances underlying unexpected outcomes. Examples include among others: detecting behaviour change by analyzing user profiles,  and identifying the causes of some anomalies such as bank frauds by analyzing customer interactions.
In this paper, we demonstrate the effectiveness of our approach in a real-life medical application: the elderly fall-prevention and showcase how the resulted explanation AND-OR trees can be used to give reliable recommendations to physicians and assist decision-makers. Our approach runs in a polynomial time and gives systematically and substantially good results.
\end{abstract}

\section{Introduction}
\label{Introduction}

New Information Technology and Communications, such as E$-$health and E-commerce, expert systems, intelligent agents, are spreading throughout the world. To provide the required services, most of these real-world systems rely on a massive amount of data. Information is commonly obtained from a variety of sources and is frequently uncertain and unreliable. Information uncertainty stems from, but is not limited to, three reasons: (1) imprecision (unreliable acquisition sources); (2) obsolescence (out-of-date information); and (3) incompleteness (limited or missing information).
This study focuses on the uncertainties surrounding information obsolescence.

Information obsolescence may be caused by, but not restricted to, information aging or by the acquisition of a new observation on a specific subject that \emph{contradicts} what we currently know about this subject, given a representation model, or both. For instance, a person's age-related decline in vision is an example of information aging, which results in obsolete information about a person's vision. An example of obsolescence due to the occurrence of a contradiction between observations is as follows:
Consider a person who is in good health and drives her car regularly. Now, you learn that she has a serious disease that causes the loss of eyesight (new information). Since a blind person cannot drive a car (general knowledge), the new piece of information prompts us to update the older one (she drives her car regularly) since it clearly became obsolete after the arrival of the new observation.

In this paper, we focus on information obsolescence related to the contradiction between newly acquired observations and existing ones given a representation model. We address the new challenge that concerns the design of an Obsolete Information Recommendation System (OIRS). Our OIRS' goal is to continuously track and monitor observations, detect contradictions between observations, and recommend all the observations that may become obsolete and require updating.

Our approach is primarily based on studying causal dependencies between the given observations in order to gain a better understanding, comprehension, and insight into the occurrence of some events, namely the contradiction. Indeed, it explains the causes of  the occurrence of a contradiction between given observations (why do we have a contradiction?), as well as the different alternatives to remove this contradiction (what are the possible observations responsible for contradiction? What if we update the values of some observations?). Furthermore, it is intended to be applied in an uncertain environment characterized by a lack of information as we may not have observations on all the  characteristics of a given subject. For these different reasons, we choose to use a Causal Bayesian network (CBN) to represent our knowledge.
The overall architecture of our OIRS is shown in Fig.~\ref{fig1:agent}. We will use a running example of a consistent set of information related to an elderly person to illustrate how it works:

\begin{figure}[H]
    \begin{center}
        \includegraphics[scale=0.4]{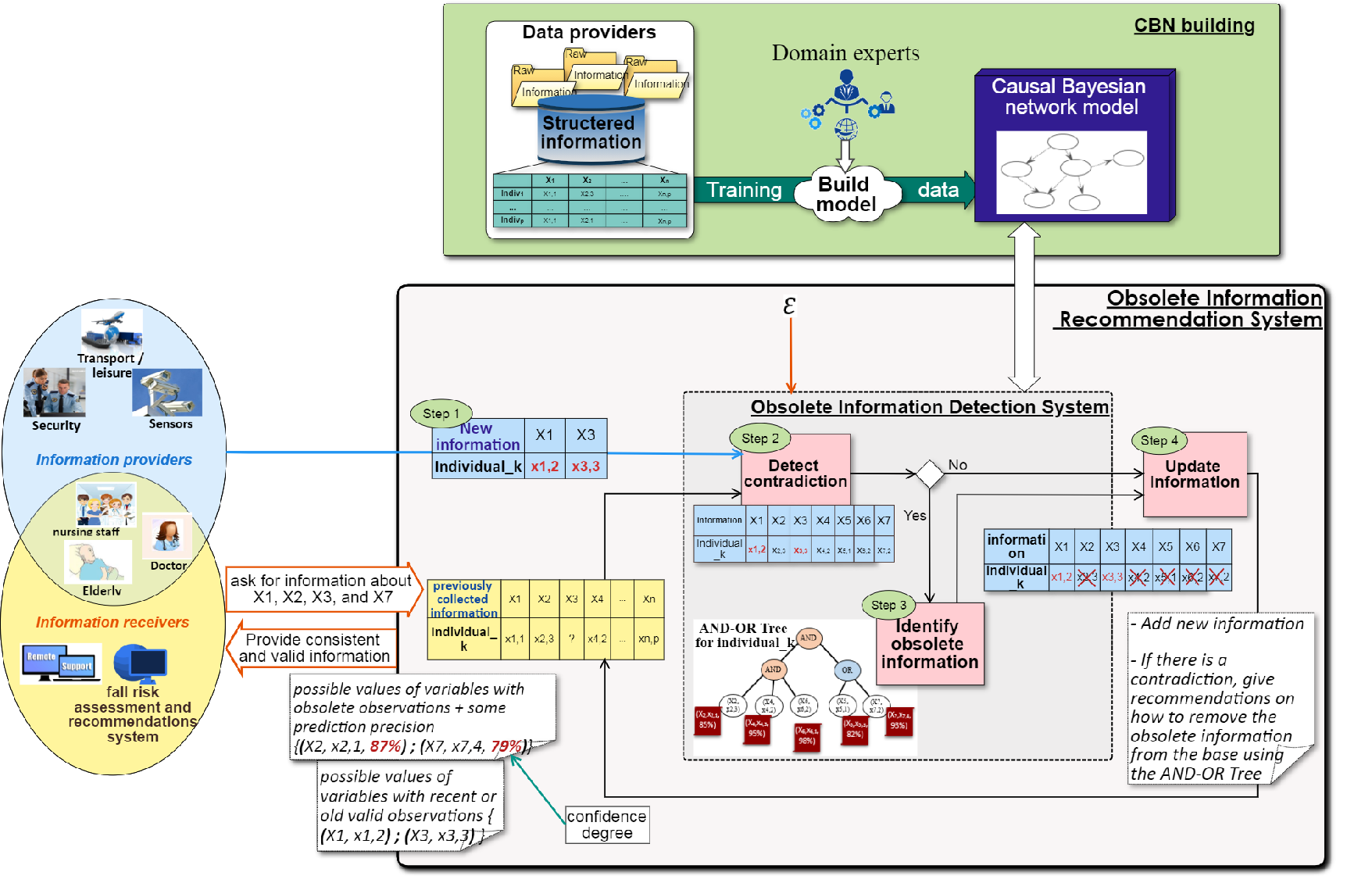}
    \end{center}
\caption{Obsolete Information Recommendation System}
\label{fig1:agent}
\end{figure}

\begin{example}
Mrs. Wilson has a good walking ability, drives her car, and goes shopping regularly. Her house is equipped with a sensor network that detects her movements and these sensors are in good operating condition.
\end{example}
\begin{sloppypar}
This is the old information that we know, acquired from information providers, and stored in the elderly database row reserved for Mrs. Wilson. By inference (using a CBN approved by experts), she regularly leaves her home.  In step 1, new observations are recorded. For our elderly person, new information, that is supposed to be certain, is given by a sensor indicates that:
\emph{Mrs. Wilson did not leave her home during the last 15 days}. In step 2, using a CBN model and based on a given threshold $\epsilon$, contradictions between the newly arriving observations and the existing ones are detected. For our elderly person, the new observation raises a contradiction since it is no longer consistent with the older ones given the CBN. Next, in step 3, we identify possible obsolete information. 
In our running example, our system detects that either Mrs. Wilson \emph{did not leave home}, \textbf{or} \emph{sensor network has broken down}.
Finally, in step 4, we provide recommendations for experts on how to update obsolete observations. So in our example, we recommend the responsible for monitoring to start by checking the sensor's state.
After checking the sensor's state, a new observation stating that the sensor network works well arrives, and we start a new cycle. This second new piece of information (captured in step 1) affirms contradiction (in step 2). Our Information Update System at step 3, thus, should deduce that Mrs. Wilson's behavior has changed as she has not actually left home during the last 15 days. So, she did not drive and shop as she used to. Then, the information \emph{she drives her car} \textbf{and} \emph{she does shopping regularly} have become obsolete to the new observation and must be updated. Our system  recommends, at step 4, the removal of obsolete information from the database row reserved for Mrs. Wilson with a certain priority. In order to assist decision-makers (i.e., information receivers), we assume that older information is more likely to be obsolete, furthermore, we recommend the most probable substitute values that can replace the obsolete ones.
Our system can, therefore, supply one of the specialist doctors of the elderly with additional information, helping to better understand her health status. 
Finally, our recommendations are associated with a confidence level in order to manage uncertainty.

\end{sloppypar}

The main focus of this paper is on steps 2 and 3 of the OIRS, i.e., the Obsolete Information Detection System (OIDS).
In this work, we assume that recently acquired information is \emph{reliable} and \emph{consistent} with each other.
Indeed, a piece of information, already stored in the database, may turn out to be obsolete because of the occurrence of a new event, considered as certain, that contradicts it, given the CBN.
We delegate the construction of the CBN to a separate module given by the green zone above the OIRS in Fig.~\ref{fig1:agent}.

Though closely related to belief revision theory \cite{agm1985}, we do not update the knowledge base (CBN) but only update personal information about a single subject. 
Other works,  such as \cite{chennovel}, propose a method to understand the abnormal trajectories of a pedestrian by using a sparse representation-based classification model.
entails analyzing a pedestrian's path and determining whether or not it is normal behavior based on training samples. Despite the fact that our method is based on the detection of abnormal behavior, it is not a straightforward problem. Rather, it entails identifying the obsolete information that is responsible for the anomalies and updating it in order to restore the personal database's consistency. 
To the best of our knowledge, our work tackles a novel problem that has not yet been addressed in the literature.
In this paper, we  demonstrate the applicability of the  proposed approach in the context of a European elderly fall-prevention project.

This paper is organised as follows. In Sect.~\ref{section2}, we describe formal background and notation. Sect.~\ref{section3} is an overview of related works. 
In Sect.~\ref{section4}, we  propose  a  new  concept regarding $\epsilon$-Contradiction among observations and study its theoretical properties. In Sect.~\ref{section5}, we give and discuss the obsolete information detection algorithm and its complexity. In Sect.~\ref{section6}, we show how to calculate the optimal threshold beyond which contradictions are detectable. In Sect.~\ref{section7}, we present our empirical study on a real-life application example: the elderly fall-prevention context. 
Finally, we conclude in Sect.~\ref{section8} and outline our future research directions.

\section{Formal background}
\label{section2}

Unless otherwise noted, throughout this paper, we denote elderly database attributes with uppercase letters such as $X_1$, $X_2$, $X_3$. The domain of an attribute $X_i$ is denoted with $\mathcal{D}(X_i)$. Specific values, also called observations, taken by those attributes are denoted with lowercase letters $x_{1,1}$, $x_{2,1}$, $x_{3,2}$ with $x_{i,j} \in \mathcal{D}(X_i)$, $j \in \{1, ..., \#\mathcal{D}(X_i)\}$. In this paper, we are only concerned with finite domains.

In this work, we choose to use a causal Bayesian network \cite{pearlbayesian,pearl2018book,luo2019causal} to represent our knowledge. Several facts justify this choice: first, we work within an uncertain environment characterized by a lack of information, as we may not have observations on all the characteristics of an elderly person. CBNs are thus a powerful probabilistic model that allows reasoning with incomplete data.
Second, our approach is primarily based on studying dependencies and causal relationships between the given observations to provide us with an understanding, comprehension, and insight into the occurrence of some events. It can also be used to formalize, measure, and deal with different unfairness (contradictory) scenarios underlying a dataset.
Indeed, it explains how and why an attribute influences other variables in a dataset, the causes of  the occurrence of a contradiction between given observations (why do we have a contradiction?), as well as the different alternatives to remove this contradiction (what are the possible observations responsible for contradiction? What if we update the values of some observations?). So, CBN represents a flexible useful tool in this respect, as it can be used to formalize, measure, and deal with different unfairness (contradictory) scenarios underlying a dataset.
Third, in real life, the newly acquired information may be uncertain, contrary to what we have supposed. So CBN proves to be very relevant for managing such a type of information through likelihood evidence \cite{pearl:plaus} and fixed and non-fixed probabilistic observations \cite{bloemeke:agent,mdpuncertain}.
Fourth, once obsolete observations are detected, our system can recommend values that can replace the obsolete ones to help the user make the right decision. This recommendation can be carried out using an intervention-based strategy that recommends the possible interventions on how to substitute the obsolete values. A highly accurate prediction model alone is not able to provide guidance when reasoning what might happen if we take an action, that is, intervene by changing the values of some variables supposed to be obsolete. Consequently, CBN proves to be very relevant to predict the result of external interventions on a variable via \emph{causal inference} \cite{pearl:causal}.

A CBN (also known as a Markovian model) provides a general modeling framework for representing complex causal networks and can be used to model different causal queries, including inferences about observations and interventions.  A CBN is a Bayesian network (BN) \cite{pearlbayesian,jensenintro,darwichemodeling} where the parents of each vertex are its direct causes. 
The direct causes of $X_i$ are the variables that will change the distribution of $X_i$ as we vary them, as we perfectly intervene in the entire system. A perfect intervention on some $X_i$ is an independent cause
of $X_i$ that sets it to a particular value, all other things remain equal.
Operationally, this just wipes out all edges into $X_i$ and  make it a constant. All other things remain equal.

As with the Bayesian network (BN), each CBN is associated with a pair $(G, \Theta)$, where $G = (\textbf{X}, \textbf{E})$ is a Directed Acyclic Graph (DAG) with nodes $\textbf{X} = \{X_1,...,X_n\}$ corresponding to the database attributes, and directed edges (also called arcs) $\textbf{E}$ that connect these nodes and denoting \emph{causal influence}. Indeed, given two variables $X_1$ and $X_2$, we say that $X_1$ precedes $X_2$ causally if experimental interventions that change the value of $X_1$ can affect the distribution of $X_2$ but not vice versa.
The second component of the pair, namely $\Theta$, represents a set of parameters that quantifies the network and is stored as a set of conditional probability tables (CPT) for each node.
A CBN over the set of attributes $\textbf{X}$ uniquely defines a joint probability distribution given by the chain rule~\ref{eq1}. It is defined by the product of the local probability distributions on each variable $X_i$, $P(X_i \mid Pa(X_i))$, where $Pa(X_i)$ denotes the set of the parents of the node $X_i$ in $G$. So, $X_j$ is parent of $X_i$ if there is a directed link from $X_j$ to $X_i$ (i.e., $(X_j, X_i) \in \textbf{E}$). $X_i$ is then called child of $X_j$. 

\begin{equation}
    P (\textbf{X}) = \prod_{i=1}^{n} P (X_i \mid \emph{Pa}(X_i))
    \label{eq1}
\end{equation}

We say that variables $X_i$ and $X_j$ are conditionally independent given a set of variables $\textbf{Y} \subset \textbf{X}$, if $P(X_i | X_j, \textbf{Y}) = P(X_i | \textbf{Y})$.

In this work, we consider the CBN as a stable component and its construction and updating is not part of the aim of this paper. The CBN is passed as a parameter to our detection module. It is used only to represent causal dependencies between observations and to compute the conditional probability of one node, given values assigned to the other attributes. 

In order to restrict the obsolete observations search set, we will later use the idea of a Markov boundary of a node $X_i$ in a BN and the concept of \emph{active path} that creates dependencies between nodes. These concepts also remain valid when we approach CBN.

The \emph{Markov boundary} \cite{pearlbayesian} of a node $X_i$ is a subset of nodes that "prevents" $X_i$ from being affected by any node outside the boundary.
One of $X_i$'s Markov boundaries is its Markov blanket. 

The \emph{Markov blanket}, MB($X_i$), of a node $X_i$ is unique and comprises $X_i$'s parents (direct causes), $X_i$'s children (direct effects), and $X_i$'s spouses
(i.e., other parents of the node’s children) in the CBN.
So with complete information, when computing the conditional probability of a node $X_i$, MB($X_i$) forms a natural feature selection, as all features outside the Markov blanket can be safely deleted from the CBN. This can often produce a much smaller CBN without compromising the computing accuracy.

The \emph{active path} is defined based on the concept of d-separation in CBNs \cite{pearl:probabilistic,pearlbayesian}. Indeed, given two variables $X_1$ and $X_k$ from the CBN, a path $X_1 - X_2 - ... - X_k$ from $X_1$ to $X_k$ is active if for each consecutive triplet in this path one of the following conditions is true:
\begin{itemize}
\item[-] $X_{i-1} \to X_i \to X_{i+1}$, and $X_i$ is not observed,

\item[-] $X_{i-1} \leftarrow X_i \leftarrow X_{i+1}$, and $X_i$ is not observed,

\item[-] $X_{i-1} \leftarrow X_i \to X_{i+1}$, and $X_i$ is not observed,

\item[-] $X_{i-1} \to X_i \leftarrow X_{i+1}$, and $X_i$ is observed or one of its descendants.
\end{itemize}

We denote by $N$ the number of CBN's variables and by $E$ the number of CBN's edges. Sets are denoted by boldface capital letters.
We define some sets and functions, which we will use extensively throughout this paper:
\begin{itemize}

\item $\textbf{OBS}= \{(X_i, x_{i,j}), i \in I, I \subseteq [1, N], x_{i,j} \in \mathcal{D}(X_i), j \in \{1, .., \#\mathcal{D}(X_i)\} \}$ the set of pairs representing a variable and its observed value, relating to a single individual of the database.

\item $\emph{Parents} (X_i):   X_i \mapsto \{(X_j, x_{j,k}), (X_j, X_i) \in \emph{\textbf{E}}, k \in \{1, .., \#\mathcal{D}(X_j)\}\}$, a function that returns the set of $X_i$ ’parents and their associated observations.

\item $\emph{Children} (X_i): X_i \mapsto \{(X_j, x_{j,k}), (X_i, X_j) \in \emph{\textbf{E}}, k \in \{1, .., \#\mathcal{D}(X_j)\}\}$, a function that returns the set of $X_i$ ’children and their associated observations.
\begin{sloppypar}
\item $ \emph{Spouses} (X_i): X_i \mapsto \{(X_j, x_{j,k}), \emph{Children} (X_i) \cap \emph{Children} (X_j) \neq \emptyset, k \in \{1, .., \#\mathcal{D}(X_j)\} \}$, a function that returns the set of $X_i$ ’spouses and their associated observations if they exist.
\end{sloppypar}
\end{itemize}

\begin{sloppypar}
Let $o_{new}$ denote the newly observed value of a variable $O_{new} \in \textbf{X}$ such that $(O_{new}, o_{new})$ was not in $\textbf{OBS}$ at the previous iteration. 
When there was a previous observation on $O_{new}$, the new one replaces it, meaning that $(O_{new}, o_{new})$ replaces the previous element related to $O_{new}$ in $\textbf{OBS}$. When no previous observation on the variable $O_{new}$ was present in $\textbf{OBS}$, the element  $(O_{new}, o_{new})$ is added to the set $\textbf{OBS}$. Thus, we define the set $\textbf{OBS'}$ as follows:
\begin{itemize}
\item $\textbf{OBS'}= \textbf{OBS} \setminus \{(O_{new}, o_{new})\}$. 
\end{itemize}

\end{sloppypar}

In order to simplify, we assume that $\textbf{OBS}$ is initially consistent, i.e. it does not contain contradictory observations until the acquisition of $o_{new}$ — or $\textbf{OBS'}$ is consistent. We assume also that newly acquired observations are reliable and consistent with each other.

Table~\ref{tabnotation} provides a description of the  parameters and notation used throughout this paper.

\begin{table}[h]
    \centering
    \begin{tabular}{|M{1.5cm}|p{8.5cm}|}
         \hline
         \emph{\textbf{Notation}} & \emph{\textbf{Description}} \\
         \hline
          $O_{new}$ & newly observed variable of the CBN \\
         \hline
          $o_{new}$ & newly observed value on the variable $O_{new}$\\
         \hline
         $\textbf{OBS}$ & set of all observed variables and its observed values  relating to a single individual of the database \\
         \hline
         $\textbf{OBS'}$ & set of all observed variables and its observed values  relating to a single individual of the database, deprived of $(O_{new}, o_{new})$  \\
         \hline
         $G$ & Directed acyclic graph \\
         \hline
         $\theta$ & set of parameters that quantifies the CBN \\
         \hline
         $\textbf{X}$ & set of variables of the CBN \\
         \hline
         $\textbf{E}$ & set of directed edges that connect $\textbf{X}$ in the CBN \\
         \hline
         $N$ & the number of variables in the CBN \\
         \hline
         $E$ & the number of directed edges in the CBN \\
         \hline
         $Pa(X_i)$ & set of parents of a variable $X_i$ in the CBN \\
         \hline
         $\epsilon$ & threshold that represents the confidence level of having a contradiction between some observations, $0 \leq \epsilon \leq 1$ \\
         \hline
         $\textbf{S}_{o_{new}}$ & set of all obsolete observations, $\textbf{S}_{o_{new}} \subseteq \textbf{OBS'}$ \\
         \hline
         $N_d$ & the size of the set $\textbf{S}_{o_{new}}$ \\
         \hline
         $MB(X_i)$ & Markov Blanket of a variable $X_i$ \\
         \hline
         $\textbf{S}_i$ & subset of $\textbf{S}_{o_{new}}$ that contains dependent variables and their observed values \\
         \hline
         $\textbf{S}_i^{AND}$ & set containing each observation in $\textbf{S}_i$ that is individually $\epsilon$-Contradictory to $o_{new}$, given the CBN \\
         \hline
         $\textbf{S}_i^{OR}$ & set containing each observation of $\textbf{S}_i$ that is not individually $\epsilon$-Contradictory to $o_{new}$, given the CBN \\
         \hline
         $N_s$ & the number of observations in $\textbf{S}_i$ \\
         \hline
         $\mathcal{T}$ & Explanation AND-OR Tree \\
         \hline
         $\mathcal{B}$ & a CBN \\
         \hline
         $N_s$ & the number of observations in $\textbf{S}_i$\\
         \hline
         $\mathcal{S}$ & a test set of contradictory and non contradictory scenarios labeled by experts \\
         \hline
         $\textbf{C}_i$ & a scenario example represented by the pair of newly acquired information accompanied by a sequence of pairs of some previously acquired observations \\
         \hline
         $c_i$ & label attributed to $\textbf{C}_i$ by experts,  $c_i = 1$ if $\textbf{C}_i$ is declared contradictory by the experts, $c_i = 0$ otherwise \\
         \hline
    \end{tabular}
    \caption{Summary of notation}
    \label{tabnotation}
\end{table}


\section{Related works}
\label{section3}

The term \emph{obsolescence} was introduced in 1820-1830, but became widely known to people in 1932 in the United States thanks to the book by Bernard London \cite{londonending}. The interpretation of the obsolescence of an object remains vague and varied and can be defined by the fact of becoming obsolete, out of date, outdated, invalid, etc.
A  bibliographic search shows that, to the best of our knowledge, there are no works that tackle information obsolescence as defined in our work.
Nevertheless, there are some related works in the information revision field in general. We can broadly classify the existing approaches into two major categories: those which address the problem of the knowledge-base revision, and those which concern the updating of an information base.

The first category falls within the theory of belief revision  \cite{agm1985,jiang2017modified,deng2015generalized}. This area has been treated from several points of view: symbolic versus numerical as well as logical versus probabilistic. Since the logical environment does not take into account explicit measures of uncertainty, some other revision approaches have used the frameworks of both possibility and probability theories. \cite{gyenis:modal,darwiche1997logic,baioletti:l1}.

The Bayesian belief revision \cite{strossner:compositionality,brown:modal} is a particular type of probabilistic belief revision, in which a belief about the target variable of the graph is represented by its conditional probability given the evidence that has been observed. In the Bayesian context, as with belief change in general, the two processes \emph{revision} and \emph{update} have to be distinguished: first, \textit{belief updating} (also called probabilistic inference) consists of calculating $P(X|\textbf{Y})$, the posterior probabilities of target nodes $X$, given some observed values of evidence nodes $\textbf{Y}$. Second, \textit{belief revision}, also called Bayesian belief revision, aims to change the initial probability distribution $P$ on variables $\textbf{X}$ of the CBN in the bases of a local distribution $R(X)$, $X \in \textbf{X}$. The same methods used in probabilistic belief revision apply to Bayesian belief revision.

Although our approach fits into the information revision context, it does not address the same problem as defined above in belief revision theory. Indeed, all the revision methods mentioned above are intended to revise the knowledge by changing the representation model in response to changes in the domain. However, in our work, we are not talking about knowledge revision. The knowledge is supposed to be stable. It is represented by a CBN that describes some characteristics of a given subject, i.e., we assume that, during an agreed period, the network structure for the graphical model and the CPTs are an accurate representation of the knowledge about the domain and therefore it does not change. The primary object of our work is, therefore, to update a database that contains information as observations on CBN variables upon the acquisition of new observations that contradict them, without changing the CBN. 

The second category includes methods of managing the information base obsolescence. The existing approaches can be broadly categorized into two types: information update based on its aging over time, and information update based on tracing the output errors back to the input data.

The first approach stems from studying the information age and its evolution.
Information aging has acquired a specific interest in the domain of web information. The spread of events and information in social networks and news pages is a particular focus in this domain. For example, \emph{aging theory} is applied to represent variations in the number of publications related to a topic or event \cite{chenlife,paikparameterized}. In these work, decay functions are used to model temporal decreases. Decay functions are also used in the domain of sensor data to manage the aggregation of data with different ages \cite{cormodetime}. Indeed, the older the data, the less weight it has in the aggregated data summary.

In \cite{lpt:segment}, authors describe a data-driven methodology for the automatic identification of text segments in encyclopedic resources, which contain information requiring updating named "obsolescence segments". They defined an obsolescence segment as a text that is likely to have changed between the time of the publication of the article and the moment of its reading. It thus needs to be updated. Wang \emph{et al.} propose in \cite{wzl:datacollection} a partial coding of a sensor network which solves the problem of removing obsolete information in coded data segments to accommodate newly collected one. In that work, the information obsolescence process is mainly based on the buffer size of the sensors and the age of the information. 
In \cite{cdmg}, the authors proposed specific functions "\emph{expiry functions}" that attribute a confidence degree to each observation, considering some parameters: acquisition date, validity period, observation type, update frequency, history, etc. 
In \cite{hxjh:rule}, authors resolve the problem of entity resolution (also known as duplication detection, record linkage) on temporal data. According to them, certain attributes of records referring to the same real-world entity in a database may change over time. All of these records may be valid and proper for describing a certain entity only at a particular time period. So they develop a rule-conducted uniform framework for resolving temporal records by integrating data quality rules.

In industry, obsolescence of a component or a system is the fact that this system is no longer useful, simply because of technical developments, the impossibility of maintenance, an unaffordable cost or that the product is no longer available for purchase in its original form from the original manufacturer or producer \cite{mellalobsolescence,sanguriforecasting}. In \cite{sandborndata}, authors propose a data mining based approach to electronic part obsolescence forecasting. The proposed approach is based on forecasting electronic part vendor-specific windows of obsolescence using historical last-order or last-ship dates. In \cite{grichiapproach}, authors propose a stochastic method for predicting the product life cycle in order to help companies improving obsolescence forecasting and reduce its impact in the supply chain. The proposed method is based on the simulation of demand data using Markov chain and homogeneous compound Poisson process.
M. Mastrangelo \emph{et al.} propose in \cite{mastrangelorisk} a Weibull-based conditional probability method to predict microelectronic component obsolescence.
In \cite{UsabilityFirst}, authors propose an approach to manage inventory obsolescence to improve retail performance. They design an Obsolescence Mark Down stock model to quantify stock aging accurately and appropriately. They argue that not all merchandise categories age at the same rate in terms of loss of saleable value and therefore think of creating a different aging profile for each category of merchandise.

In \cite{rens:stochastic}, the author design a framework with which
an agent can deal with uncertainty about its observations. This framework
includes how to integrate 'expired' and 'prevalent' observations into the agent's beliefs. In that work, each observation has a meaningful period for which it can be thought of as certainly true, and the author studies the veracity of information by attaching an 'expiry date' to each observation. In \cite{farazifundamental}, authors study the \emph{age of information} in a general multi-source multi-hop wireless network with explicit channel contention. An algorithm to generate near-optimal periodic status update schedules based on sequential optimal flooding is developed. In \cite{wuoptimal}, authors use the metric \emph{Age of Information}, the time that has elapsed since the last received update was generated, to measure the \emph{freshness} of the status information in a network. 

These studies are close to ours, except that in these cases, information obsolescence is defined regarding its aging over time rather than its inconsistency with newly acquired information. So, an information acquired at time \emph{t} may no longer be reliable at time \emph{t+1}, i.e., it can become obsolete. In these work, authors use a time-stamped data acquired over a time interval. Thus, the goal is to track the change of observations over time and identify those that have become obsolete. However, in our work, the obsolescence of information is conditioned, in addition to its age, by the arrival of a new event that is assumed to be valid, which contradicts it and therefore renders it invalid.

The second approach falls within the theory of causality and explanation in databases \cite{livshits:shapley,likr:principles,labreuche:explaining,meliou:causality}. It consists of understanding the underlying causes of a particular observation by determining the relative contribution of features in machine-learning predictions \cite{labreuche:explaining}, the responsibility of tuples to database queries \cite{bertossi:data,lbks:shapley}, or the reliability of data sources \cite{cpt:using}.
In \cite{livshits:shapley}, a Shapley value is used to quantify the extent to which the database violates a set of integrity constraints.
It consists of assigning to individual tuples a level of responsibility to the overall inconsistency, and thereby prioritize tuples in the explanation or inspection of dirt. In that work, the authors carried out a systematic investigation of the complexity of the Shapley value in common inconsistency measures for functional-dependency violations.
In \cite{mgns:tracing}, authors focus on determining the causes of a set of unexpected results, possibly conditioned on some prior knowledge of the correctness of another set of results.

In those works, the main idea is to observe some variables, make the inference, get the predicted values, and evaluate the resulted values based on some prior knowledge of the correctness of another set of results. If errors are detected, then try to find among the observed variables those responsible for this dire prediction. The problem here differs from ours for many reasons. First, we are not tackling a classification/prediction problem. Our approach is a bit complicated as obsolete observations are given in a particular form, an explanation AND-OR tree, and therefore it can not be viewed as a simple classifier learning problem.
Second, in functional dependencies-based works \cite{livshits:shapley}, systems detect errors in the output data, i.e., after the input data are integrated and propagated, then trace the output errors back to the input data. In order to check the correctness of the outputs, authors have access to several output items that they know are correct. However, in our work, errors are detected earlier, before the input data is transformed and integrated. Third, in those works, the inputs are a set of Boolean/numerical variables, and the treatment is based on propositional formulas and fits into the logical environment, which does not consider explicit measures of uncertainty. However, in real life, the input data is often uncertain, and our method can perfectly fit this constraint through uncertain evidence in the CBN, as they represent a powerful model for knowledge representation and reasoning under uncertainty. Moreover, in Conditioned Causality-based work \cite{mgns:tracing}, the application can often detect such errors from user feedback or based on the user’s subsequent actions and reactions to the provided recommendations (target data). However, in our method, the contradiction detection is fully automatic and does not require user intervention. And last, but not least, in those works, the complexity of the proposed methods remains a worrying issue as it can introduce exponential blow-ups in the size of the handled logical expression. However, in our work, we propose a quadratic-time approach to handle obsolescence.
Table~\ref{tab1} highlights the main differences between existing work and our approach.

\begin{table}[t]

\begin{tabular}{|p{1.8cm}|M{1.9cm}|M{2cm}|M{1.8cm}|M{2cm}|}
\hline
& Revise Bayesian belief in presence of new information \cite{gyenis:modal,schwering:belief}  & Update information when detecting output errors \cite{mgns:tracing,livshits:shapley}  & Update information based on its aging over time \cite{lpt:segment,cdmg,sandborndata}  & Update information when it contradicts  other new acquired one (Our approach) \\ 

\hline
Knowledge representation   & Bayes’ rule / probability distributions /  propositional formula & Functional dependency constraints /  propositional formula & Association rules / probability distributions /  expiry functions & Causal Bayesian network  \\ 

\hline
Information representation & Propositional information   & Propositional information   & Propositional sentence  & Observations on the CBN variables   \\ 
\hline

Check the information aging & $\times$  &  $\times$ & \checkmark &  \checkmark \\ 

\hline
Detect inconsistencies  & \checkmark & \checkmark & $\times$  & \checkmark  \\ 

\hline
Support Incomplete information & $\times$ & $\times$ & $\times$ &  \checkmark \\ 

\hline
Support uncertain information & $\times$ & $\times$ & $\times$ & \checkmark \\ 

\hline
Update the knowledge-base  & \checkmark & $\times$ & $\times$ & $\times$ \\ 

\hline
Update the information base & $\times$ & \checkmark & \checkmark & \checkmark \\ 

\hline
Output  & New probability distributions & Set of source data responsible for bad prediction  & Expired information  & AND-OR explanation tree of obsolete information  \\ 
\hline
\end{tabular}
\caption{A comparative summary table that highlights the main differences between existing work and our approach}
    \label{tab1}

\end{table}

\section{Approximate contradiction}
\label{section4}
Recall that so far we assume that we have a consistent database, a CBN and that a recently acquired information $o_{new}$ on a variable $O_{new}$ related to a specific subject is certain.

A contradiction occurs when the new observation $o_{new}$ added to the data-base row relating to this subject, is no longer consistent with the rest of the observations about this subject (denoted by $\textbf{OBS'}$), given the CBN model that represents the dependencies among variables. Hence, theoretically, a contradiction between observations occurs when the conditional probability of the new observation given other observations is zero:

\[
P (O_{new} = o_{new} | \textbf{OBS'}) = 0
\]

However, since the CBN is a probabilistic knowledge representation model, there is always a degree of uncertainty related to the inference we draw from it. Hence, in practice, we shall not expect the conditional probabilities to be exactly zero but rather very close to 0. For that reason, we introduce a contradiction probability tolerance value $\epsilon$ to reflect those uncertainties related to the probabilistic dependencies among the variables in the CBN\footnote{We show later how to carefully choose the threshold $\epsilon$ associated with the given CBN}. The approximate contradiction is thus defined as follows:

\begin{Definition}[\textbf{${\epsilon}$-Contradiction}]
\label{definition1}
Given a CBN, a set of observed variables $\textbf{OBS'}$, a new observation $o_{new}$ on a variable $O_{new}$, and a real number $0 \leq \epsilon \leq 1$.
$\textbf{OBS'}$ is ${\epsilon}$-Contradictory to $o_{new}$ when 
\[
P (O_{new} = o_{new}|\textbf{OBS'})
\leq \epsilon
\]
\end{Definition}

So, we say that we have an $\epsilon$-Contradiction when there is a subset of $\textbf{OBS'}$ of observations that have become obsolete and contradict $o_{new}$. We refer to the set of all obsolete observations as $\textbf{S}_{o_{new}}$. 
In the following, a set of observations is said to be consistent when there is no $\epsilon$-Contradiction with $o_{new}$.
At this stage, we can introduce the following proposition.

\begin{Proposition}
\label{proposition1}
Given a CBN, a new observation $o_{new}$ on a variable $O_{new}$ and a set of observed variables $\textbf{OBS'}$.
If $\textbf{OBS'}$ is $\epsilon$-Contradictory to $o_{new}$, then there exists a unique subset $\textbf{S}_{o_{new}} \subseteq \textbf{OBS'}$ of obsolete observations that is $\epsilon$-Contradictory to $o_{new}$ and such that $\textbf{OBS'} \setminus \textbf{S}_{o_{new}}$ is consistent, i.e., $P (O_{new} = o_{new}|\textbf{OBS'}\setminus \textbf{S}_{o_{new}}) > \epsilon$.
\end{Proposition}

\begin{proof}\emph{\textbf{Existence.}}
~ \\ If $\textbf{OBS'}$ is $\epsilon$-Contradictory after acquiring $o_{new}$, then we consider two hypotheses: either $o_{new}$ is uncertain or invalid and then contradicts $\textbf{OBS'}$, or there are, among the observations of $\textbf{OBS'}$, those that have become obsolete and then contradict $o_{new}$. Since we assumed at the beginning of the paper that the newly acquired observation is certain, then the first hypothesis will be discarded. Hence there exists a subset of $\textbf{OBS'}$ that contains obsolete observations responsible for the contradiction.
\end{proof}

\begin{proof}\emph{\textbf{Uniqueness.}}
~ \\ Assume that there exists two different sets of obsolete observations $\textbf{S}_1 \subseteq \textbf{OBS'}$ and $\textbf{S}_2 \subseteq \textbf{OBS'}$ that are $\epsilon$-Contradictory to $o_{new}$ and such that 
$P (O_{new} = o_{new}| \textbf{OBS'}\setminus\textbf{S}_1) > \epsilon$ and $P (O_{new} = o_{new}| \textbf{OBS'}\setminus\textbf{S}_2) > \epsilon$.
Since $\textbf{S}_2 \subseteq \textbf{OBS'}$, then the withdrawal of $\textbf{S}_1$ from $\textbf{OBS'}$ is not sufficient to restore its consistency with $o_{new}$, i.e., $P (O_{new} = o_{new}|\textbf{OBS'}\setminus \textbf{S}_1) \leq \epsilon$. This contradicts the assumption that  $P (O_{new} = o_{new}| \textbf{OBS'}\setminus\textbf{S}_1) > \epsilon$. So $\textbf{S}_{o_{new}}$ is unique.
\end{proof}

In what follows, we show how to progressively obtain the set $\textbf{S}_{o_{new}}$ from $\textbf{OBS'}$, looking for all possible obsolete observations.

\subsection{Contradictory set restriction}

\label{pruning}

\begin{figure}[t] 
  \centering
  \includegraphics[scale=0.26]{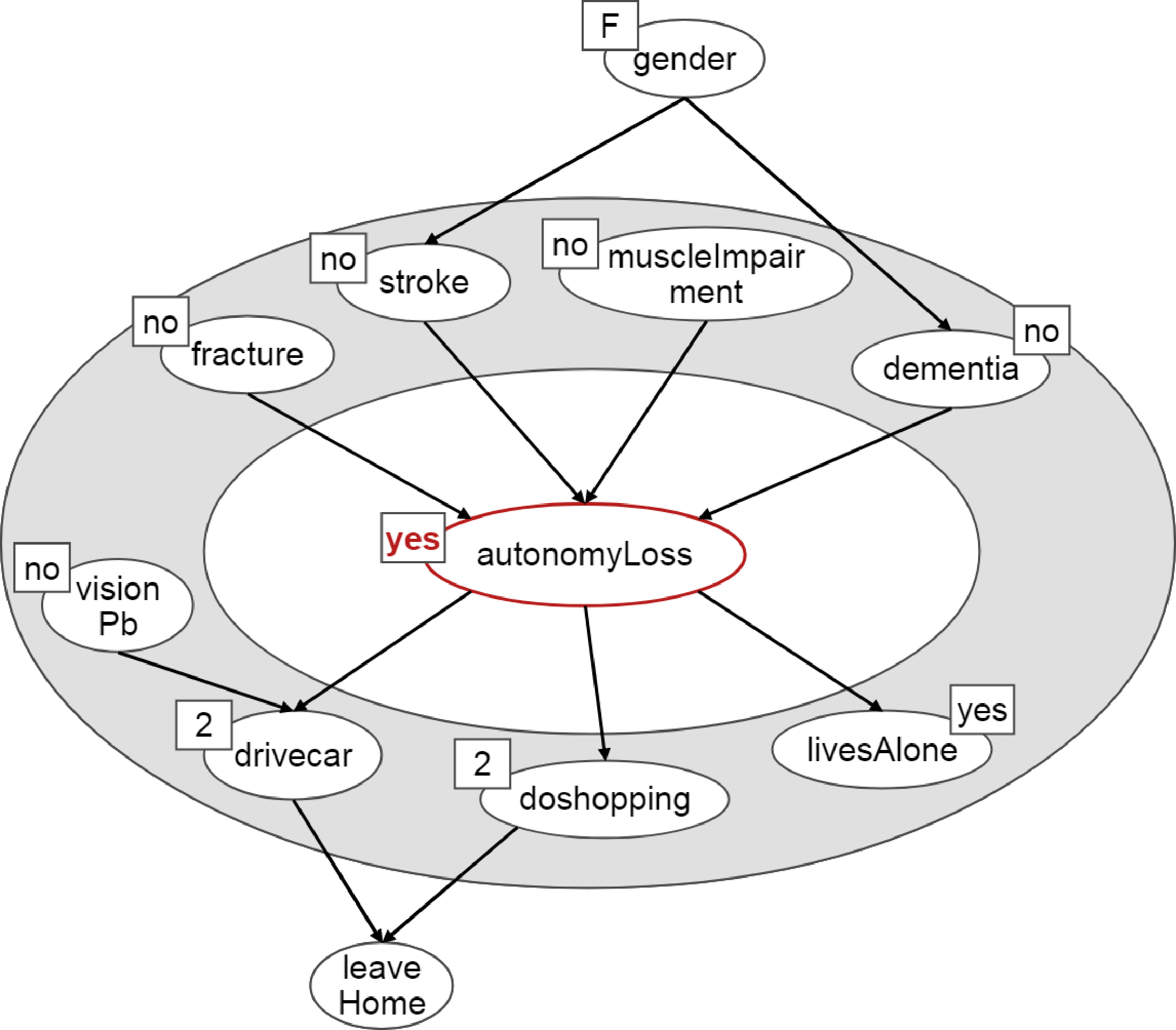}
  \caption{The Markov Blanket of $O_{new}$ illustrated on a sample scenario on part of the CBN}
  \label{exp}
\end{figure}

When $\textbf{OBS'}$ is $\epsilon$-Contradictory to $o_{new}$, it does not mean that all observations in $\textbf{OBS'}$
are obsolete. Let us consider the example shown in Fig.~\ref{exp} on which we approach the steps of the obsolete information detection process.

\begin{example}
\label{example2}
At time $t_1$, the set of observations $\textbf{OBS}$ includes nine observations related to an elderly person: she is a woman, she lives alone and she does not have a stroke, does not have dementia, does not have muscle impairment,and she does not suffer from fracture. Moreover, she has a good eyesight, drives her car and does her shopping regularly. At time $t_2$, new information showing that Mrs. Wilson has lost her autonomy arrives.

This new information contradicts the set $\textbf{OBS'}$, but, it is clear that the contradiction occurs regardless of the elderly’ sex. Hence, information about the person’s sex is not a necessary part of the set $\textbf{S}_{o_{new}}$ of obsolete observations.
\end{example}

According to the Bayesian net assumption, when all nodes of the Markov Blanket of the node $O_{new}$ are observed, $O_{new}$ becomes independent of any other node. We now consider that case, \textit{i.e.}, we assume now that the spouses, parents and children of $O_{new}$ are all observed (see Fig.~\ref{exp}). Thus all the variables of MB($O_{new}$) have an observation in $\textbf{OBS'}$. 
The gray area in Fig.~\ref{exp} represents the Markov blanket as explained in Sect.~\ref{section2}. Here, we can restrict $\textbf{OBS'}$ to the subset $\textbf{S}_{o_{new}}$ of observations included in MB($O_{new}$) as follows:

\[
\textbf{S}_{o_{new}} \subseteq  \textbf{OBS'} \cap MB(O_{new})
\]

Therefore, when we have a contradiction, instead of exploring the whole set $\textbf{OBS'}$ looking for obsolete observations, we only need to check observations on variables of the Markov Blanket of $O_{new}$.

Now, if we do not consider the assumption about the parent, children, and spouse nodes of the new observation, the set $\textbf{OBS'}$ of contradictory observations can be restricted in $\textbf{S}_{o_{new}}$ as follows instead:
\[
\textbf{S}_{o_{new}} \subseteq  \textbf{OBS'} \textit{ such that there exists an active path between }
\]
\[ O_{new} \textit{ and each element of } \textbf{S}_{o_{new}}. \]

We conclude this section with the following proposition:

\begin{Proposition}
\label{proposition2}
Given a CBN and a new observation $o_{new}$ on a variable $O_{new}$, a set $\textbf{OBS'}$ that is $\epsilon$-Contradictory to $o_{new}$ and a non-empty subset $\textbf{S}_{o_{new}} \subseteq \textbf{OBS'}$ such that $\textbf{S}_{o_{new}}$ contains all the nodes of $\textbf{OBS'}$ for which there exists an active path with $O_{new}$, the following statements are true:
\\
(1) Any observation of $\textbf{S}_{o_{new}}$ may belong to the set of obsolete observation after acquiring
$o_{new}$; and
\\
(2) Any observation of $\textbf{OBS'} \setminus \textbf{S}_{o_{new}}$ does not belongs to the set of obsolete observations after acquiring $o_{new}$.

\end{Proposition}

\begin{proof}
This proposition comes from the fact that (1) any variable $X$ in $\textbf{S}_{o_{new}}$ is dependent on $O_{new}$ through the existence of an active path from $X$ to $O_{new}$ and hence may be obsolete; (2) Any variable in $\textbf{OBS'} \setminus \textbf{S}_{o_{new}}$ is independent of $O_{new}$ by using the notion of active path and hence cannot be obsolete.
\end{proof}

\subsection{Contradictory set decomposition}
\label{decomposition}
Up till now, we have shown that $\textbf{OBS'}$ can be restricted to $O_{new}$'s dependent variables in $\textbf{S}_{o_{new}}$. The decomposition phase takes place in two stages: as a first step, we decompose $\textbf{S}_{o_{new}}$ further into subsets, $\textbf{S}_i$, by bringing the observations on dependent variables together. 
The decomposition is based on studying the causal relationships between variables. Indeed, if two variables are conditionally dependent, updating one variable can influence the other. This allows us to interpret the interaction between the given observations and give us a more precise idea of the priority of updating them.

\begin{figure}[t]
  \centering
  \includegraphics[scale=0.8]{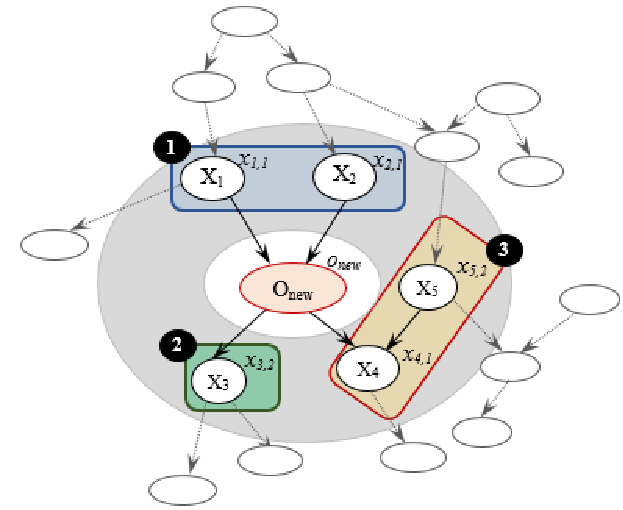}
  \caption{An example of decomposition}
  \label{decompose}
\end{figure}

Depending on the causal structure of the CBN, the subset $\textbf{S}_i$ can take one of the following three forms: as shown in Fig.~\ref{decompose}:
\begin{itemize}
    \item $\textbf{S}_i$ contains all the causes of the common effect $O_{new}$,
    \item $\textbf{S}_i$ only contains an effect of $O_{new}$, when there are no other causes of this effect, and
    \item $\textbf{S}_i$ contains an effect of $O_{new}$ and the other variables that cause this effect if they exist.
\end{itemize}

Let us consider the example shown in Fig.~\ref{exp} in which  $\textbf{S}_{o_{new}}$ is in the grey region. Indeed, once $O_{new}$ is observed, its direct causes $fracture$, $stroke$, $muscleImpairment$, and $dementia$ are dependent on one another, unlike its direct effects $driveCar$, $doShopping$, and $livesAlone$ which are conditionally independent. So we gather all direct causes of $O_{new}$ in the same subset and each effect with its causes, except $O_{new}$, in a subset.
The result of such decomposition give the set $\textbf{S}_{o_{new}} = \{ \{$\emph{(fracture,no), (stroke,no), (muscleImpairment,no), (dementia,no)}$\}$, $\{$\emph{(visionPb,no), (driveCar,2)}$\}$, $\{$\emph{(doShopping,2)}$\}$, $\{$\emph{(livesAlone,yes)}$\}\}$.

In the general case, we decompose $\textbf{S}_{o_{new}}$ by bringing the observations on dependent variables together. Indeed, as mentioned earlier in Sect.~\ref{pruning}, two variables are dependent in the CBN if and only if there exists an active path between them.
In some cases, when decomposing observations, subsets may overlap. The same observation may appear concurrently in different subsets $\textbf{S}_i$, as it may depend on several subsets of observations. However, we show later that this specific case does not raise any problems.

As a second step, we explore all subsets $\textbf{S}_i$ resulting from the first phase of decomposition, and for each one, we check whether it is consistent with $o_{new}$ or not. If so, it will be ignored and will not be considered in the rest of the obsolete information identifying process.
This aims to reduce the search space of obsolete observations.
So as a result of these two stages of decomposition, we have a set $\textbf{S}_{o_{new}}$ of subsets $\textbf{S}_i$ of observations and their related variables. Each subset obeys these two decomposition properties:

\begin{Property}
\label{property1}
Each subset $\textbf{S}_i$ resulting from the decomposition phase contains dependent variables.
\end{Property}

\begin{Property}
\label{property2}
Each subset $\textbf{S}_i$ resulting from the decomposition phase is $\epsilon$-Contradictory to $o_{new}$, given the CBN.
\end{Property}

At this point, the main obsolete information detection process will revolve around each $\epsilon$-Contradictory subset $\textbf{S}_i$. Following the example shown in Fig.~\ref{decompose}, and respecting the 3 possible forms of the subset  $\textbf{S}_i$, the observations into $\textbf{S}_i$ are organized so that in case of contradiction, three explanations are possible: 
\\

\textbf{\emph{Explanation 1}}: in the case where  $\textbf{S}_i$ contains all causes of $o_{new}$ (group 1 in Fig.~\ref{decompose}), at least  $x_{1,1}$ or $x_{2,1}$ or both cannot cause $o_{new}$. In such a case, we distinguish two aspects of each observation $x_{i,j}$ contained in the set $\textbf{S}_i$.
Either $x_{i,j}$ is individually $\epsilon$-Contradictory to $o_{new}$ (i.e., $P(o_{new}|x_{i,j}) \leq \epsilon$) or not. Thereby, we can decompose this $\textbf{S}_i$ further into two disjoint $\epsilon$-Contradictory subsets:
\begin{itemize}
    \item The \emph{AND-Set}, $\textbf{S}^{AND}_{i}$, containing each observation in $\textbf{S}_i$ that is individually $\epsilon$-Contradictory to $o_{new}$, given the CBN,
    
    \item The \emph{OR-Set}, $\textbf{S}^{OR}_{i}$, containing each observation in $\textbf{S}_i$ that is not individually $\epsilon$-Contradictory to $o_{new}$, given the CBN, but the entire \emph{OR-Set} is $\epsilon$-Contradictory to $o_{new}$, given the CBN.
\end{itemize}

All the observations in the \emph{AND-Set} are obsolete and need to be updated, which is not always the case for those in the \emph{OR-Set}. Let's explain this tricky situation with the following example:
if we suppose that leaving home ($O_{new}$) is conditioned by both events: driving a car ($X_1$) or shopping ($X_2$). We know that an older adult does not drive his car \emph{($X_1$, no)} and does not go shopping \emph{($X_2$, no)}. A new observation given by an external sensor showing that he has left his house \emph{($O_{new}$, yes)}. Certainly, this new observation raises a contradiction with what we already know, i.e., $\{(X_1, no), (X_2, no)\}$ is $\epsilon$-Contradictory to $(O_{new}, yes)$.
However, if we inspect each of these observations separately, we find that each of them is not individually $\epsilon$-Contradictory to $(O_{new}, yes)$.

Indeed, since these observations are conditionally dependent,  $P(O_{new}= yes | X_2 = no)>\epsilon$ comes from the fact that the elderly may have driven her car. The same applies to $P(O_{new}= yes | X_1 = no) > \epsilon$. So, both \emph{($X_1$, yes)} and \emph{($X_1$, yes)} are placed in the \emph{OR-Set}, which meaning that updating one of these two old observations is enough to remove the $\epsilon$-Contradiction. However, with the available knowledge at our disposal, we are not able to accurately infer which one(s) should be updated.
Consider this scenario now replacing the observations \emph{(no, no, yes)} respectively on the variables ($X_1$, $X_2$, $O_{new}$) with \emph{(yes, yes, no)}. The observations on the two variables $X_1$ and $X_2$ are individually $\epsilon$-Contradictory to \emph{($O_{new}$, no)} and will be placed into the \emph{AND-Set}.
\\

\textbf{\emph{Explanation 2}}: in the case where  $\textbf{S}_i$ contains the effect of $o_{new}$ (group 2 in Fig.~\ref{decompose}), the contradiction is explained by the fact that $o_{new}$ cannot cause $x_{3,2}$, i.e., that $x_{3,2}$ is individually $\epsilon$-Contradictory to $o_{new}$ and is therefore classified in an \emph{AND-Set} of $\textbf{S}_i$.
\\

\textbf{\emph{Explanation 3}}: in the case where  $\textbf{S}_i$ contains the effect of $o_{new}$ and its other causes (group 3 in Fig.~\ref{decompose}), the contradiction is explained by the fact that at least $o_{new}$ or $x_{5,2}$ or both cannot cause $x_{4,1}$. As we previously assumed that the newly acquired observation $o_{new}$ is certain, meaning it cannot be objected, then the contradiction is explained by the fact that:

\begin{itemize}
    \item either $x_{5,2}$ cannot cause $x_{4,1}$ given $o_{new}$ and then need to be updated, or $x_{4,1}$ cannot be the effect of both $x_{5,2}$ and $o_{new}$ and then need to be updated. In this case, both $x_{5,2}$ and $x_{4,1}$ are not individually $\epsilon$-Contradictory to $o_{new}$ and then are classified in an \emph{OR-Set} of $\textbf{S}_i$. An example is ($O_{new}$, $X_4$, $X_5$) = (\emph{leaveHome, NbOfExitsGPS, GPSState}), ($o_{new}$, $x_{4,1}$, $x_{5,2}$) = (\emph{yes, 0, OK}).
    
    \item $x_{4,1}$ cannot be the effect of $o_{new}$ independently of $x_{5,2}$, meaning that $x_{4,1}$ is individually $\epsilon$-Contradictory to $o_{new}$ and then is classified in an \emph{AND-Set}. An example is ($O_{new}$, $X_4$, $X_5$) = (\emph{diabetes, drugsNb, cardiovascularDrugs}), ($o_{new}$, $x_{4,1}$, $x_{5,2}$) = (\emph{yes, 0, no}).
\end{itemize}  

Note that in general, the \emph{OR-Set} may contain extra-elements, i.e., observations that are not part of the $\epsilon$-Contradiction. In such a case, we proceed by elimination. We remove an observation from this set, and we check whether the remaining set is $\epsilon$-Contradictory to $o_{new}$ given the CBN. If so, then the observation removed is not among those responsible for contradiction and it must be then ignored. We apply the same treatment to all observations of the \emph{OR-Set} in order to have a set containing only potentially obsolete observations, and such that the withdrawal of any observation from this set restores its consistency with $o_{new}$.

\subsection{AND-OR tree composition}
\label{composition}

As a result of the decomposition phase, we obtain the set $\textbf{S}_{o_{new}}$ of obsolete observations, which satisfies the proposition~\ref{proposition1}.

Each $\textbf{S}_i \in \textbf{S}_{o_{new}}$ is divided into two subsets: $\textbf{S}^{AND}_{i}$ and $\textbf{S}^{OR}_{i}$. The main aim of the current phase is to combine the results of these subsets to create an explanation AND-OR tree  whose internal nodes are labeled either \emph{AND} or \emph{OR} and whose leaves represent all possible obsolete observations in each $\textbf{S}^{AND}_{i}$ and $\textbf{S}^{OR}_{i}$ subset.

The explanation AND-OR tree is constructed as follows. We introduce a root node labeled \emph{AND}. Then, for each subset $\textbf{S}_i$, we introduce an \emph{AND} node whose parent is the root node. Next, for each \emph{AND-Set} (resp. \emph{OR-Set}) of $\textbf{S}_i$, we introduce an \emph{AND} (resp. \emph{OR}) node whose parent is the corresponding node of $\textbf{S}_i$ and a child leaf node for each observation in $\textbf{S}^{AND}_{i}$ (resp. $\textbf{S}^{OR}_{i}$). Each leaf node is labeled with the obsolete observation. 

\begin{figure}[t]
  \centering
  \includegraphics[scale=0.52]{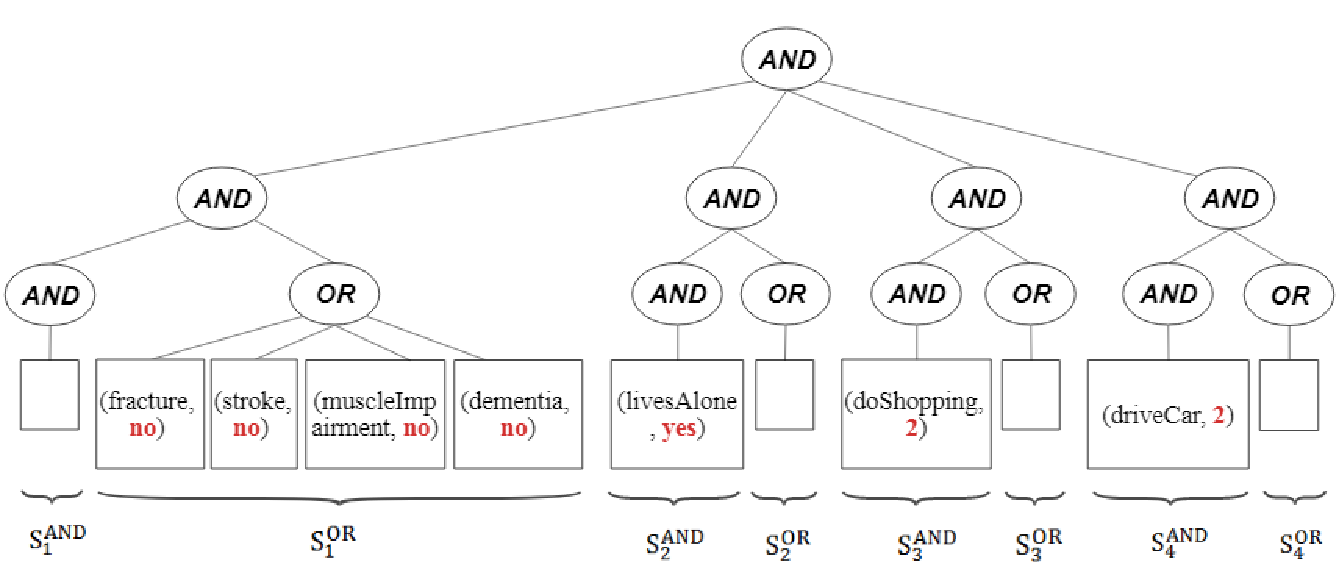}
  \caption{The explanation AND-OR tree related to example~\ref{example2}}
  \label{fig5:Tree}
\end{figure}

\begin{figure}[t]
  \centering
  \includegraphics[scale=0.39]{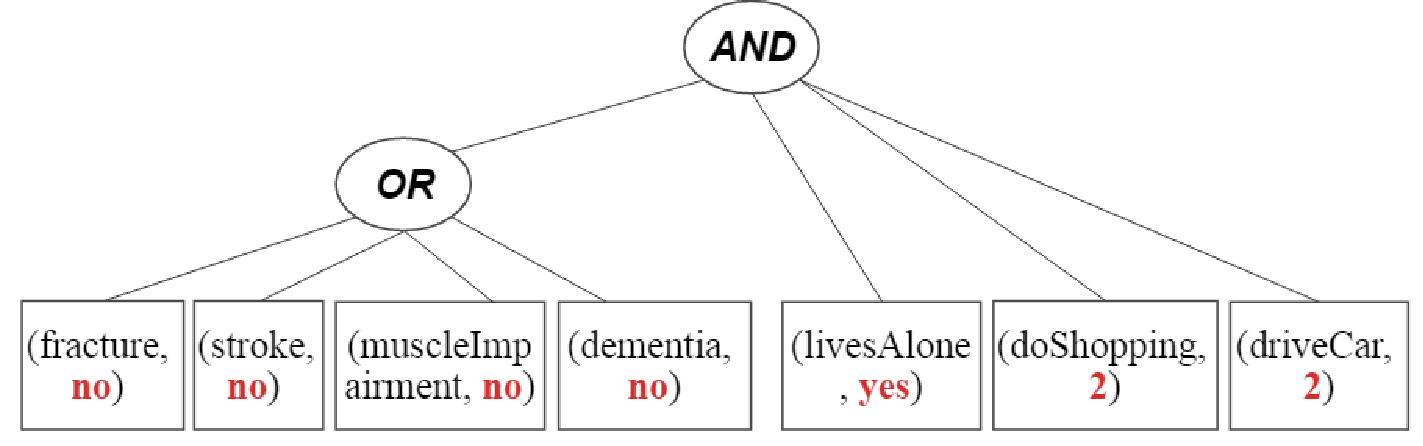}
  \caption{The explanation AND-OR tree related to example~\ref{example2} after being simplified}
  \label{fig6:tree}
\end{figure}

Fig.~\ref{fig5:Tree} shows the resulting explanation AND-OR tree associated with Example~\ref{example2}.
The left branch of the tree corresponds to the $\epsilon$-Contradictory Subset $\textbf{S}_1$, which is divided into two disjoint $\epsilon$-Contradictory Subsets the \emph{AND-Set}: $\textbf{S}_1^{AND}$ = $\emptyset$, and the \emph{OR-Set}: $\textbf{S}_1^{OR}$ = \{(fracture, no), (strokeTIA,no), (muscleImpairment, no), (dementia, no)\}. The same reasoning applies for the other three sets $\textbf{S}_2$, $\textbf{S}_3$, and $\textbf{S}_4$.
The resulting AND-OR tree can be simplified as shown in Fig.~\ref{fig6:tree}.

The AND-OR tree represents precisely the set of obsolete observations $\textbf{S}_{o_{new}}$ and explains the logical relationships among its \emph{AND-Set}s and \emph{OR-Set}s. 
Indeed, given a new observation $o_{new}$ and an AND-OR tree $\mathcal{T}$ that represents the set of obsolete observations relative to $o_{new}$, the following three propositions are true:

\begin{Proposition}
\label{proposition3}
For each observations $x \in \textbf{OBS'}$, $x \notin \mathcal{T}$ if and only if $x$ is not obsolete.
\end{Proposition}
\begin{proof}

"$\Longrightarrow$" if $x \notin \mathcal{T}$ then $x$ is not part of the set of obsolete observations since $\mathcal{T}$ reflects this set. So $x$ is not obsolete. \\
"$\Longleftarrow$" $x$ is not obsolete means that $x$ is not part of the set of obsolete observations and since $\mathcal{T}$ represents this set then $x \notin \mathcal{T}$.
\end{proof}

\begin{Proposition}
\label{proposition4}
All observations of the \emph{AND-Set} are obsolete.
\end{Proposition}
\begin{proof}

Follows immediately from the fact that each observation in the \emph{AND-Set} is individually $\epsilon$-Contradictory to the new observation given the CBN.
\end{proof}

\begin{Proposition}
\label{proposition5}
At least one observation of the \emph{OR-Set} is obsolete.
\end{Proposition}

\begin{proof}
Follows immediately from: (1) the fact that none of the observations is individually $\epsilon$-Contradictory to the new observation given the CBN; (2) the \emph{OR-Set} is $\epsilon$-Contradictory to the new observation; and (3) removing any observation from the \emph{OR-Set} restores its consistency with $o_{new}$, but we can not know exactly which one is obsolete. Thus, each of the observations contained in the \emph{OR-Set} is likely to be involved in the contradiction
\end{proof}

As stated previously in Sect.~\ref{decomposition}, a same observation may appear concurrently in different subsets $\textbf{S}_i$. However, it behaves in precisely the same way in all the subsets to which it belongs. Indeed, if this observation is individually $\epsilon$-Contradictory (resp. $\epsilon$-non Contradictory) to $o_{new}$ given the CBN, it will appear in the \emph{AND-Set} (resp. \emph{OR-Set}) of each $\textbf{S}_i$ and in any case, it must (resp. may) be updated. This is, therefore, a simple duplication that can easily be handled.

\section{Obsolete Information Detection Algorithm}
\label{section5}
We now define the main steps of the algorithm~\ref{alg:algorithm} for building an AND-OR tree. 
\begin{algorithm}[h]
\caption{Obsolete Information Detection Algorithm (OIDA)}
\begin{algorithmic}[1]
\label{alg:algorithm}
\renewcommand{\algorithmicrequire}{\textbf{Input:}}
\renewcommand{\algorithmicensure}{\textbf{Output:}}
\REQUIRE  $(O_{new}, o_{new})$, $\mathcal{B}$
\ENSURE Explanation AND-OR Tree \\
\hspace{-0.58cm}\textbf{Parameters}: $\epsilon$: a real number, $0 \leq \epsilon \leq 1$\\

\STATE let $\textbf{OBS'}$ be the set of observations and their associated variables in $\mathcal{B}$ except $(O_{new}, o_{new})$.

\STATE $\textbf{S}_{o_{new}} = Prune (\textbf{OBS'}, O_{new}, \mathcal{B})$.

\IF {$IsContradictory (\textbf{S}_{o_{new}}, (O_{new}, o_{new}), \mathcal{B}, \epsilon)$}

\STATE $\textbf{S}_{o_{new}} = Decompose (\textbf{S}_{o_{new}}, \mathcal{B}, \epsilon)$.

\STATE let $\textbf{S}_{o_{new}}=\{\textbf{S}_1, ..., \textbf{S}_i, ..., \textbf{S}_p\}$ such that each $\textbf{S}_i$ is $\epsilon$-Contradictory to $o_{new}$ given $\mathcal{B}$.

\STATE AND-OR-Tree $\leftarrow Compose(\textbf{S}_1) \wedge ... \wedge Compose(\textbf{S}_i) \wedge ... \wedge Compose(\textbf{S}_p)$.

\STATE \textbf{return} AND-OR-Tree
\ELSE
\STATE \textbf{return} $True$
\ENDIF
\end{algorithmic}
\end{algorithm}
The inputs to Obsolete Information Detection Algorithm (OIDA) is a CBN $\mathcal{B}$ and a new information denoted by $(O_{new}, o_{new})$. As a first step, the new information arrives. In a second step, as shown in Fig.~\ref{fig1:agent}, our system tries to find the obsolete candidate variables that may be involved in a case of contradiction by computing $\textbf{S}_{o_{new}}$ using the function \emph{Prune} (line 2). This function takes as input the set of observed variables $\textbf{OBS'}$ and restricts it to $\textbf{S}_{o_{new}}$ representing only the dependent ones of $O_{new}$. So, instead of processing the entire CBN, we are only interested in variables that depend on $O_{new}$, which makes a considerable time gain especially since the \emph{Prune} function is linear in the number of directed edges and nodes in the CBN \cite{bdo:relevant}.

Then, OIDA checks if there is an $\epsilon$-Contradiction between $o_{new}$ and $\textbf{S}_{o_{new}}$ (line 3). If so, then our system has to look for obsolete observations (step 3 in Figure ~\ref{fig1:agent}). Indeed, considering the dependency relations between the CBN variables, we decompose $\textbf{S}_{o_{new}}$ (line 4) as explained in Sect.~\ref{decomposition} into subsets $\textbf{S}_i$ of dependent variables. Then, this function checks the consistency of each observation given by the set $\textbf{S}_i$ to $o_{new}$ and places it appropriately in either the \emph{AND-Set} or the \emph{OR-Set} to get the new decomposed set $\textbf{S}_{o_{new}}$. The \emph{Decompose} function takes $O(E \times N^{2})$, where $E$ is the number of directed edges in the given CBN and $N$ is the number of its variables.

Line 5 of OIDA traverses all elements of $\textbf{S}_{o_{new}}$ and for each one, we check whether it is consistent with $o_{new}$ given $\mathcal{B}$. If so, we delete it from $\textbf{S}_{o_{new}}$. Thus, $\textbf{S}_{o_{new}}$ contains only the $\epsilon$-Contradictory subsets $\textbf{S}_i$ to $o_{new}$ given $\mathcal{B}$, each including the \emph{AND-Set} and the \emph{OR-Set}.
This aims to reduce the search space of obsolete observations.

\begin{algorithm}[t]
\renewcommand{\thealgorithm}{}
\floatname{algorithm}{}
\caption{ \textbf{Function} \emph{Compose}}
\begin{algorithmic}[1]
\renewcommand{\algorithmicrequire}{\textbf{Input:}}
\renewcommand{\algorithmicensure}{\textbf{Output:}}
\REQUIRE  $\textbf{S}_i$ 
\ENSURE AND-OR sub-tree relating to $\textbf{S}_i$

\STATE Compute $\textbf{S}_i^{AND}$ the \emph{AND-Set} of $S_i$

\STATE Compute $\textbf{S}_i^{OR}$ the \emph{OR-Set} of $S_i$

\STATE let $\textbf{S}_i^{AND} = \{a_1, ..., a_k\}$

\STATE let  $\textbf{S}_i^{OR} = \{b_1, ..., b_p\}$

\IF {$ empty(\textbf{S}_i^{AND}) \wedge empty(\textbf{S}_i^{OR}$)}
\STATE \textbf{return} $True$.
\ELSE   \IF {$empty(\textbf{S}_i^{AND})$}
            \STATE \textbf{return} ($b_1 \vee ... \vee b_p$).
        \ELSE   \IF {$empty(\textbf{S}_i^{OR})$}
                     \STATE \textbf{return} ($a_1 \wedge ... \wedge a_k$).
                \ELSE
                     \STATE \textbf{return} ($a_1 \wedge ... \wedge a_k) \wedge (b_1 \vee ... \vee b_p$).
                \ENDIF
        \ENDIF
\ENDIF
\end{algorithmic}
\end{algorithm}

Line 6 of OIDA traverses all elements of $\textbf{S}_{o_{new}}$ and for each one we call the main function : $Compose(\textbf{S}_i)$. As we have explained in Sect.~\ref{composition}, the \emph{Compose} function takes as input the set $\textbf{S}_i$ subdivided into $\textbf{S}_i^{AND}$ and $\textbf{S}_i^{OR}$, and returns a sub-tree of obsolete observations relating to each $\textbf{S}_i$. This function takes $O(N_s^2)$ time complexity, where $N_s$ is the number of observations in $\textbf{S}_i$. The result of OIDA is an AND-OR Tree of all possible obsolete observations.

The OIDA maintains consistency of a database and runs in $O(N_d \times N_{s}^{2})$ where $N_d$ is the size of the set $\textbf{S}_{o_{new}}$ resulting from decomposition and $N_s$ is the number of observations in $\textbf{S}_i$. 

\begin{Proposition}
The Obsolete Information Detection Algorithm is sound and complete.
\end{Proposition}
\begin{sloppypar}
\begin{proof}\emph{\textbf{Soundness.}}
~ \\Soundness follows from proposition~\ref{proposition4} since each observation individually $\epsilon$-Contradictory to $o_{new}$, given the CBN, must be updated and from proposition~\ref{proposition5} since at least one observation of the \emph{OR-Set} must be updated.
\end{proof}
\end{sloppypar}
\begin{proof}\emph{\textbf{Completeness.}}
~ \\Completeness follows from the fact that obsolete observations are found only among $O_{new}$ dependent variables. So, we check all elements of this finite set one by one looking for obsolete ones and we return an AND-OR Tree with at least one node.
\end{proof}

The OIDA answers the following two questions:
Is there an $\epsilon$-Contradiction between the given observations? If so, what are the possible obsolete observations that cause contradiction?

The first question will be answered in the following section. Indeed, to decide if there is a contradiction or not, it is necessary to refer to a specific threshold. Sect.~\ref{section6} explains how to calculate the threshold. The second question is answered and theoretically justified  by the propositions and properties given in Sect.~\ref{section5} and will be validated experimentally in Sect.~\ref{section7}.

\section{Calculating the $\epsilon$ threshold}
\label{section6}
The contradiction detection accuracy is one of the most important criteria for the success of the obsolete information detection strategy. It is difficult to define appropriate thresholds to find a compromise between the false-positive (FP) rate (scenarios labeled as non-contradictory by the expert but $\epsilon$-Contradictory based on definition~\ref{definition1}) and the false-negative (FN) rate. In most approximation-based works, parameters are often hard to set. Various methods for parameter estimation can be proposed. In an ideal scenario, the value of $\epsilon$ can be set by a domain knowledge expert. In general, small values of $\epsilon$ are preferable. Alternatively, experimental studies and simulations on real-world databases, i.e. a data-driven approach, can be applied to choose the optimum value of $\epsilon$ and thus minimize human intervention. In what follows, an experimental design will be used to set the optimal contradiction threshold $\epsilon$, which minimizes classification error.
The main steps of the threshold calculation approach are given by algorithm~\ref{alg:algorithm2}.

The inputs of this algorithm are the CBN $\mathcal{B}$ and a database $\mathcal{S}=\{(\textbf{C}_i, c_i)\}$ of scenarios labeled by experts. Each scenario $\textbf{C}_i$ is represented by the pair of newly acquired information (variable, new observation) accompanied by a sequence of pairs of some previously acquired observations that are consistent (variable, observed value). For each scenario, $\textbf{C}_i$, a label $c_i$ given by the experts is associated such as: $c_i = 1$ if $\textbf{C}_i$ is declared contradictory by the experts, $c_i = 0$ otherwise.
Here is an example of a contradictory scenario: 

\begin{example}
$\textbf{OBS}=\{$(\emph{heartDisease, no}), (\emph{drugsNumber, 0})$\}$, the new observation is (\emph{cardiovascularDrugs, yes}) and $c_i = 1$.

The scenario $\textbf{S}_i = \{$(\emph{cardiovascularDrugs, yes}), (\emph{heartDisease, no}), (\emph{drugs Number, 0}) $\}$ is declared contradictory by the expert since $c_i=1$, and $\epsilon$-Contradictory by definition~\ref{definition1} since $P($cardiovascularDrugs=yes $|$ \textbf{OBS'})$\leq\epsilon$.
\end{example}

In this paper, our experiments were carried out using a balanced database, $\mathcal{S}_{Elderly}$, containing $560$ scenarios relating to several elderly with $280$ scenarios labeled as contradictory by the experts and $280$ scenarios labeled as non contradictory by the experts. The scenarios are of different sizes containing between $4$ and $41$ pairs (variable, observed value), which represent information about the elderly.
A part of this database ($\approx 30\%$ of the database) including $84$ contradictory scenarios and $84$ non-contradictory scenarios is used to calculate the optimal threshold associated with the CBN used. The rest of this database will be used later in Sect.~\ref{section7} to validate the second part of the OIDA (the resulting AND-OR trees).
The generation of scenarios is done using automatic processing, which consists of (1) randomly selecting variables from the given CBN, (2) assigning random observations to the selected variables, (3) arbitrarily choosing a pair (variable, observation) that represents the newly acquired information.
Scenarios are then labeled by two University Hospital physicians.

\setcounter{algorithm}{1}
\begin{algorithm}[t]
\caption{Threshold Calculation Algorithm}
\begin{algorithmic}[1]
\label{alg:algorithm2}
\renewcommand{\algorithmicrequire}{\textbf{Input:}}
\renewcommand{\algorithmicensure}{\textbf{Output:}}
\REQUIRE  database $\mathcal{S}=\{(\textbf{C}_i, c_i)\}$ of scenarios $\textbf{C}_i$ labeled by an expert, Bayesian network $\mathcal{B}$
\ENSURE threshold $\epsilon$

\FOR{each scenario $\textbf{C}_i \in \mathcal{S}$ }
\STATE let $(O_{new},o_{new})$ be the first pair of the scenario $\textbf{c}_i$ and $\textbf{OBS'}= \textbf{C}_i \setminus \{(O_{new},o_{new})\}$.
\STATE calculate $p_i = P(O_{new}=o_{new} | \textbf{OBS'})$ from $\mathcal{B}$, which represents the conditional probability associated with the scenario $\textbf{C}_i$.
\ENDFOR
\STATE Let $\textbf{CP} = \{(p_i, c_i)\}$ be the set of pairs associated with each $\textbf{S}_i$.
\STATE $\textbf{L} = CalculateFP\_FN (\textbf{CP})$ \COMMENT{\textbf{L} is a set of pairs $(FN, FP)_i$ for $12$ values of $i$ chosen in the interval $]0, 1[$.} 
\STATE \textbf{return} $OptimalThreshold(\textbf{L})$.

\end{algorithmic}
\end{algorithm}

\begin{figure}[t]
  \centering
  \includegraphics[scale=0.6]{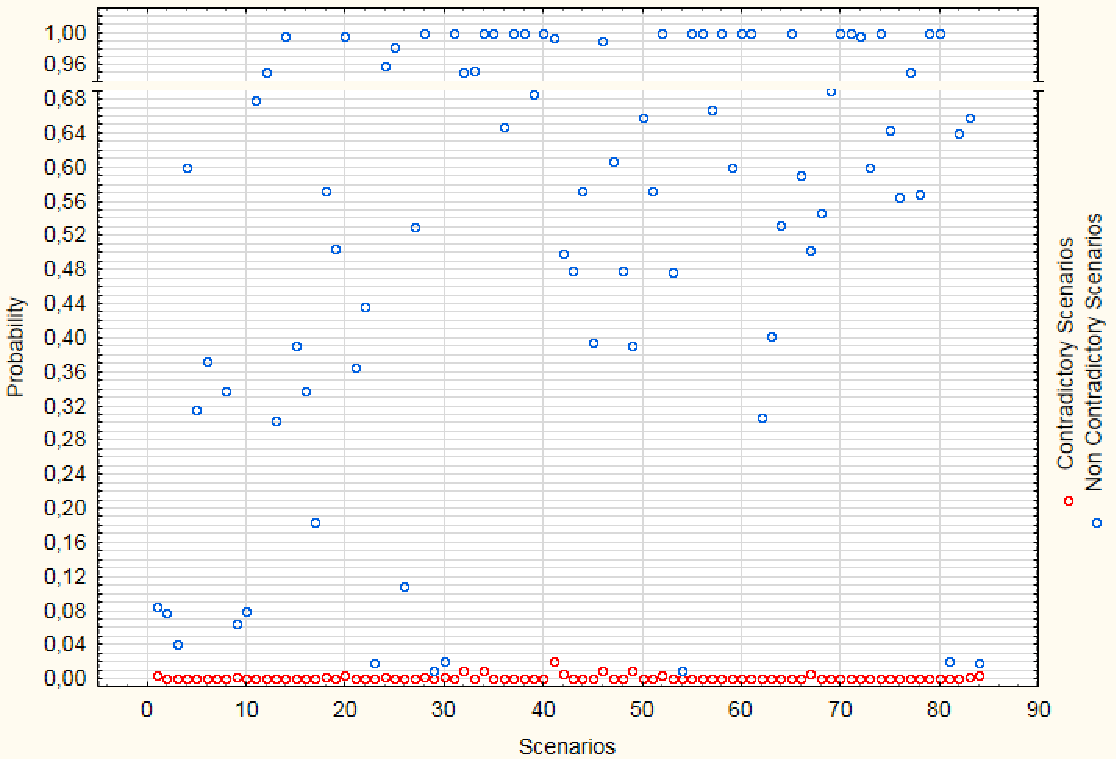}
  \caption{Probabilities of $168$ scenarios: each point corresponds to $P(O_{new}=o_{new} | \textbf{OBS'})$}
  \label{fig7:proba}
\end{figure}

To calculate the optimal threshold, we start by calculating the conditional probability for each instance in $\mathcal{S}_{Elderly}$. (line 3 of algorithm~\ref{alg:algorithm2}). This gives us a set $\textbf{CP}$ of the different conditional probabilities of $196$ scenarios. Fig.~\ref{fig7:proba} shows the overlay of the calculated probabilities.
At this stage, we notice that probabilities $p_i$ of non-contradictory scenarios vary between $0.01$ and $1$, while that of contradictory scenarios are between $10^{-11}$ and $0.02$. Since the two intervals overlap, no threshold value will completely separate them. For $\epsilon = 0$ (resp. $\epsilon = 1$), all the scenarios declared contradictory (resp. non-contradictory) by the experts are misclassified by our algorithm. So, we try to choose $\epsilon$ between $0$ and $1$ that minimizes the number of misclassified scenarios. To do so, we call the function \emph{CalculateFP\_FN}, which calculates the FP and FN rates for a sufficiently large number of thresholds ($12$ threshold values in our case) changing in small discrete steps over the entire range of the interval $]0, 1[$.

\begin{figure}[t]
  \centering
  \includegraphics[scale=1.2]{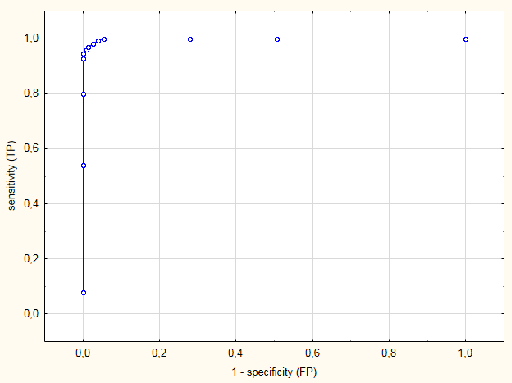}
  \caption{ROC curve for $12$ threshold values}
  \label{fig8:misclassified}
\end{figure}

The results from line 6 of the algorithm are summarized in Fig.~\ref{fig8:misclassified}. A plot of Hit Rate (along the $y$ axis) against FP rate (along the $x$ axis) for each threshold gives a Receiver Operating Characteristic curve (ROC curve). Points on the curve are got by counting the number of true and false contradictions detection and computing sensitivity and specificity at each threshold. For cases where overlap
occurs between contradictory and non-contradictory scenarios,
increasing the number of the true-positive (TP) rate will necessarily
increase the number of FP rates. The ROC curve characterizes this trade-off. 

In Fig.~\ref{fig8:misclassified}, we see that the topmost point on the left, which corresponds to the $10^{-2}$ threshold guarantees both better TP and FP rates.

\begin{figure}[t]
  \centering
  \includegraphics[scale=1.2]{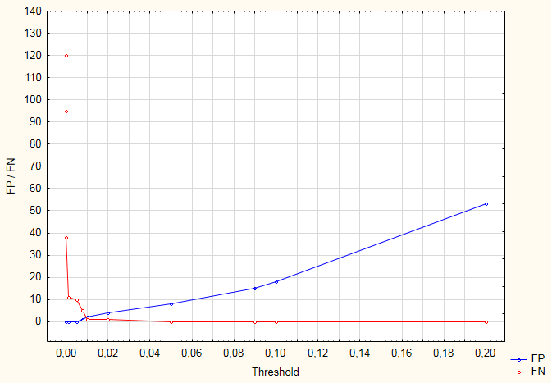}
  \caption{False-positive rate against false-negative rate}
  \label{fig10:FP_FN}
\end{figure}

To better justify the threshold value, we draw two curves as a function of FP and FN rates for the selected threshold values, as shown in Fig.~\ref{fig10:FP_FN}. It is interesting to note that the two curves intersect at the point $10^{-2}$, which corresponds to the same value designated by the ROC curve.

\section{Experimental results: towards effective fall-prevention systems}
\label{section7}

The objective of this section is to provide a first validation, after having chosen the optimal threshold, of the OIDA in a real-life application, in particular, in the elderly healthcare context, and to showcase how the resulting trees can be used to give reliable recommendations.

The sharp increase in the life expectancy of the world’s population results in a large number of older adults. This progressive aging has enormous social and economic consequences.
Meanwhile, more and more elderly are living with chronic conditions that need long-term and ongoing healthcare, preferably in their home.
As this population's age and associated chronic health conditions increase, alternatives to hospital and institutional care are needed.
Technological advances are placing increasing importance on elderly monitoring and have pushed the frontier of healthcare into the home settings \cite{malazi2018combining,liouane2018improved,raeiszadeh2019discovering}.

In this paper, we are interested in the elderly fall-prevention project.
Falls are more common in elderly people around the world and may have several painful consequences.
Consequently, in order to achieve a  reasonable degree of fall prevention among elderly,
several fall prevention strategies and tools have been proposed and tested with physicians, other health care team members, patients and some of their family members \cite{xiong2020s3d,dhiman:review}.

These systems require monitoring elderly personal database collected from hospitals, mutual health, and other associations and organizations in caring for the elderly.
This information is continually evolving and may become obsolete at a given moment and contradict other information. So, it needs to be updated in order to restore database consistency. 

In our work, the information update is carried out in two phases. First, by identifying contradictory scenarios.
Second, by identifying the possible obsolete observations that cause
contradiction in order to remove them from the database and collect additional information to expand the database.
For physicians, having such a database for each of their patients, in particular for elderly patients, can contribute to the improvement of falls prevention as well as various aging-related diseases. It is an innovative and effective way based on AI and prediction to help physicians follow their patients by providing them with some information (or predictive values in case of lack of information) at the right time for the target person. This can greatly offer improved patients care and diagnosis, reduce costs and develop relationships with their patients.

\subsection{Data description}
Our experiments are carried out using two databases: \emph{Elderly-Data} and $\mathcal{S}_{Elderly}$.
First, as part of the elderly-fall prevention project, we have access to a real-life database, \emph{Elderly-Data} that contains information on the elderly. 
It is collected during the elderly appointments with their attending physician in the University Hospital Falls Clinic of Lille over a 9-year period (2005-2014). The database includes about $1174$ patient records, each of these records was described by $435$ patient-history features (binary, denoting presence or absence of a feature or continuous, expressing the value of a feature). 
We conducted a study of these data in collaboration with hospital experts on fall prevention. This study resulted in the selection of $41$ relevant attributes associated with the characteristics of the
elderly (age, gender, BMI, etc.), the main risk factors for fall (gait/balance disorder, muscle impairment, osteoporosis, environmental risk, etc.), and possible consequences of fall (fracture, death, etc.). From this result, besides bibliographic research, and the solicitation of experts, we built and evaluated a first model of the generic knowledge embedded in a CBN denoted by $\mathcal{CBN}_{Elderly}$.
The links between variables and the parameters are pursued by asking the participated experts, including three university hospital physicians, whether they agree with them. In addition, experts evaluated the resulting marginal distributions and how the distributions change after observing certain variables and were satisfied with the results. The causal Bayesian model can, therefore, be considered valid according to four experts during a fixed time interval. For full details of the $\mathcal{CBN}_{Elderly}$ building process, we refer the author to \cite{chb:obsolete}.

\begin{figure}[ht]
\centerline{\includegraphics[scale=0.75]{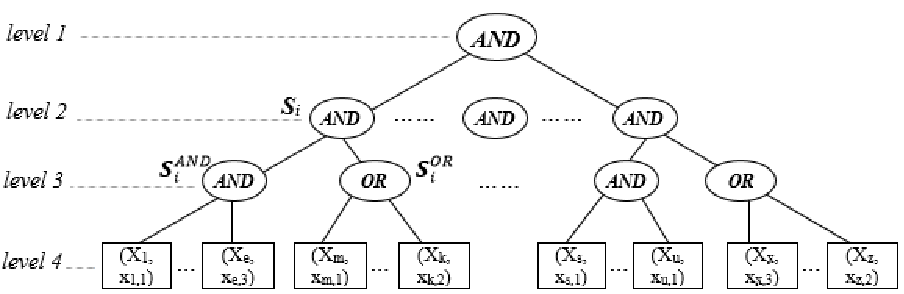}}
\caption{Hierarchical structure of the explanation AND-OR tree.}
\label{fig:tree}
\end{figure}

Then our experiments were carried out using $196$ contradictory scenarios and $196$ non-contradictory scenarios ($\approx 70\%$ of the database $\mathcal{S}_{Elderly}$ used to calculate the threshold). As we stated earlier, the generated scenarios were transmitted to two experts, an orthopedist and a neurologist, who were not involved neither in the construction of the $\mathcal{CBN}_{Elderly}$ nor in the validation of the threshold $\epsilon$. For each scenario in $\mathcal{S}_{Elderly}$, physicians established whether it is a contradictory scenario, based on their experiences. Then, for each scenario labeled as contradictory, these experts provided a list of subsets of all possible obsolete observations, such that the withdrawal of these subsets restores the consistency of the remaining observations with the newly acquired one. The resulting subsets lists were then organized into AND-OR trees following the  hierarchical structure given by Fig~\ref{fig:tree}. These trees will be compared later with the result provided by the OIDA algorithm.

\vspace{-0.5cm}
\subsection{Validation process}

The validation of our OIDS consists of two parts: measuring the accuracy of the system in detecting contradictions and evaluating the quality of the explanation AND-OR trees resulting from our OIDS.
The first part consists of measuring the false positive and the false negative rates of our system. The second part is to compare the AND-OR trees given by our system with the results given by the experts. The comparison is made at two levels. At the first level, we check the number of \emph{AND} nodes (the subsets $\textbf{S}_i$) that appear in \emph{level 2} of each tree as shown in Fig.~\ref{fig:tree}. Then, for each $\textbf{S}_i$ node, we compare the number of leaves that compose each of the $\textbf{S}_i^{AND}$ and the $\textbf{S}_i^{OR}$ nodes with those provided by the experts.

\subsection{Experimental results}

Having chosen $10^{-2}$ as the appropriate threshold in Sect.~\ref{section6}, we apply the OIDA on the $392$ remaining scenarios.
The results obtained from Step 2 of Fig.~\ref{fig1:agent} can be summarized as shown in Table~\ref{tabRes}.

\begin{table}[ht]
\begin{tabular}{lllcc}
                        &                          &                                        & \multicolumn{2}{c}{Predicted}                                        \\ \cline{4-5} 
                        &                          & \multicolumn{1}{c|}{392}               & \multicolumn{1}{l|}{$\epsilon$-Contradictory} & \multicolumn{1}{l|}{$\epsilon$-non-contradictory} \\ \cline{3-5} 
\multirow{2}{*}{Actual} & \multicolumn{1}{l|}{196} & \multicolumn{1}{l|}{Contradictory}     & \multicolumn{1}{c|}{187}               & \multicolumn{1}{c|}{9}       \\ \cline{3-5} 
                        & \multicolumn{1}{l|}{196} & \multicolumn{1}{l|}{Non contradictory} & \multicolumn{1}{c|}{15}                  & \multicolumn{1}{c|}{181}    \\ \cline{3-5} 
\end{tabular}
\caption{$10^{-2}$ Threshold contingency}
    \label{tabRes}
\end{table}

\begin{figure}[ht]
   \includegraphics[scale=1.3]{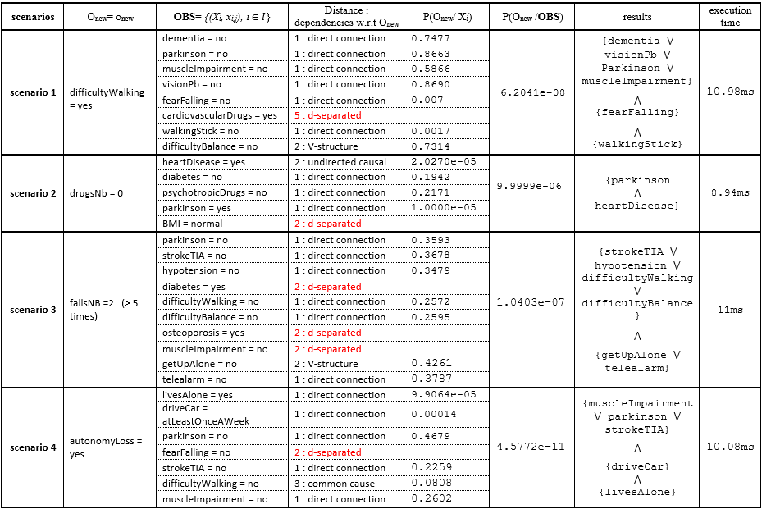}
  \caption{An example of $10^{-2}$-Contradictory scenarios processing}
  \label{fig12:results}
\end{figure}

For scenarios classified as $\epsilon$-Contradictory by our system, we apply the obsolete observation identifying process (Step 3 in Fig.~\ref{fig1:agent}). Owing to space limitations, we cannot display all the results (the AND-OR Trees) issuing from this step. However, some of the scenarios that were treated by our algorithm are shown in Fig.~\ref{fig12:results}. Column 2 (resp. 3) of the table given in Fig.~\ref{fig12:results} contains the newly (resp. previously) acquired observation(s). Column 4 refers to Sect.~\ref{pruning} in which we seek to find among the observations previously acquired those that depend on $O_{new}$. To better understand the update  process, we detail the first scenario given in Fig.~\ref{fig12:results}: the previously acquired observations set is $\textbf{OBS}= \{$\emph{(dementia,no), (parkinson,no), (muscleImpairment, no), (visionPb, no), (fearFalling, no), (cardiovascularDrugs, yes), (walkingStick, no), (difficultyBalance, no)}$\}$, the new observation is \emph{(difficultyWalking, yes)}. As shown in column 4, the variable $cardiovascularDrugs$ is independent of $O_{new}$, so it will be ignored.
The conditional probability given in column 6 means that the scenario is contradictory since the resulting value $\leq \epsilon$.

As explained in Sect.~\ref{decomposition}, the decomposition phase gives the set $\textbf{S}_{o_{new}}$ = $\{\{$\emph{(dementia, no), (parkinson,no), (muscleImpairment, no), (visionPb, no), (difficultyBalance, no)}$\}$,  $\{$ \emph{(fearFalling, no)}$\}$, $\{$\emph{(walkingStick, no)}$\}\}$.
To find out which observations in $\textbf{S}_{o_{new}}$ are part of the \emph{AND-Set} and the \emph{OR-Set}, we calculate the probability of $O_{new}$ given $X_i$ for each variable $X_i \in \textbf{S}_{o_{new}}$ dependent on $O_{new}$. In our example, we notice that the two observations $(fearFalling, no)$ and $(walkingStick, no)$ are individually $\epsilon$-Contradictory to \emph{difficultyWalking, yes)} since their associated probabilities are less than $\epsilon$.
Each observations in the first element of $\textbf{S}_{o_{new}}$ is individually $\epsilon$-non-contradictory, but the whole subset $\{$\emph{(dementia,no), (parkinson,no), (muscleImpairment, no), (visionPb, no), (difficultyBalance, no)}$\}$ is $\epsilon$-Contradictory to $o_{new}$. As explained in Sect.~\ref{decomposition}, we proceed by elimination. This process leads to ignore the observation \emph{(difficultyBalance, no)} since the withdrawal of this observation does not restore the consistency of the subset  with $o_{new}$.
At the end of treatment and as shown in column 7 of Fig.~\ref{fig12:results}, the OIDA returns the explanation AND-OR tree of possible obsolete information for scenario 1. For the sake of convenience, the tree is given by the following logical formula: $(A$ or $B$ or $C$ or $D)$ and $(E)$ and $(F)$. It means that to integrate the new observation, we must remove the two observations on $E$ and $F$, and remove either $A$ or $B$ or $C$ or $D$, depending on the user's choice.
The average execution time of a scenario is about $10$ms.

\begin{figure}[t]
  \centering
 \includegraphics[scale=0.45]{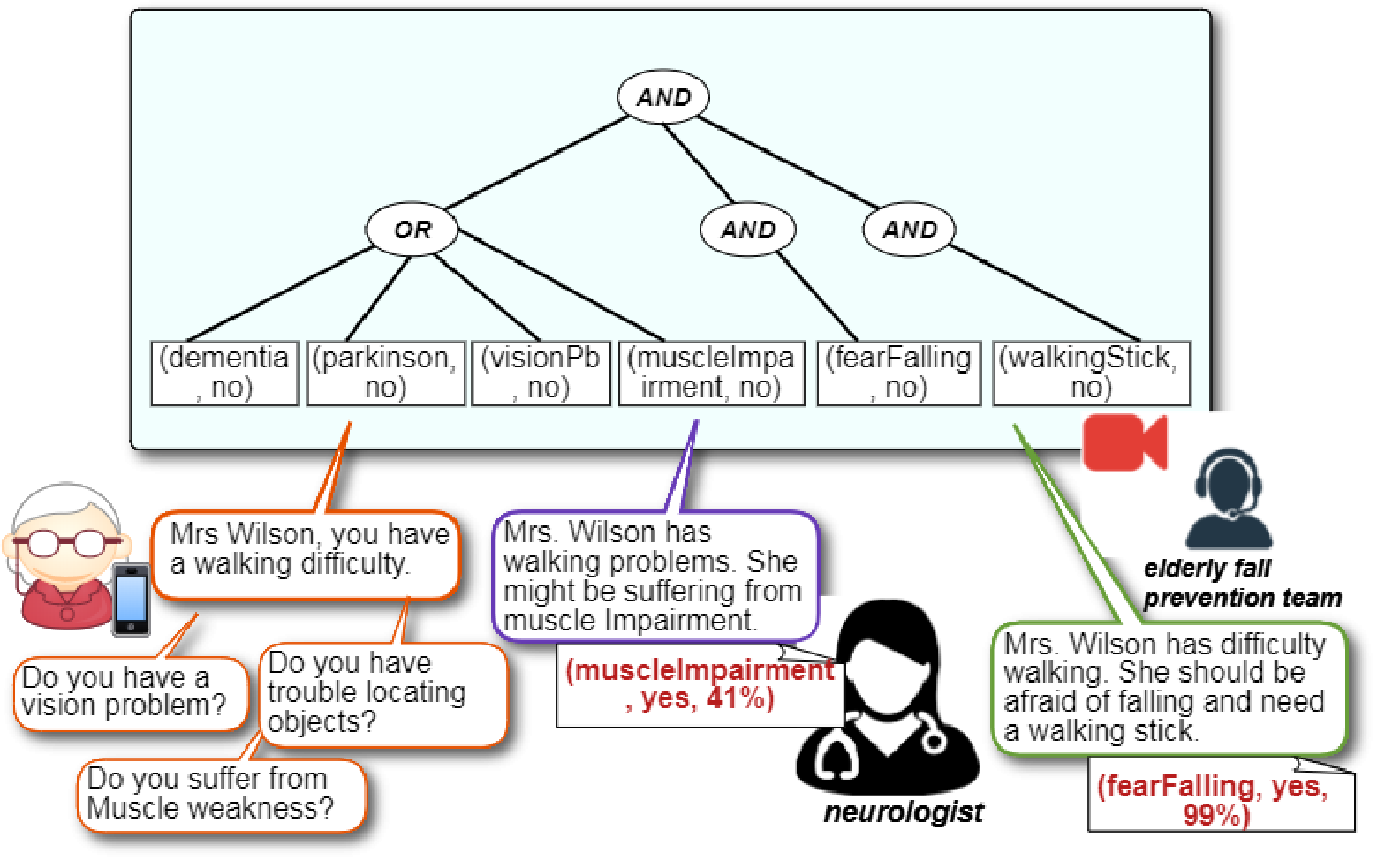}
  \caption{Personal information update  using the AND-OR Tree}
  \label{fig13:recom}
\end{figure}

For the $187$ scenarios classified by our OIDA as contradictory, the resulting AND-OR trees are ideally in line with those given by experts for $175$ scenarios. The credibility and accuracy of the resulting trees are theoretically justified by the propositions and properties given in Sect.~\ref{section5}.
Our approach efficiency is confirmed experimentally  since  our  simulations  on  a  real database  in  the  elderly  fall prevention context are very encouraging, reaching an accuracy of $93\%$.
An example of the explanation AND-OR tree use is shown in Fig.~\ref{fig13:recom}. We can use the resulting AND-OR  trees to generate a list of relevant questions to ask by physicians (or related personnel) in order to update values of obsolete observations and get the newer ones if they exist. Furthermore, we can suggest, among the observations contained in the \emph{OR-Sets}, the most likely to be updated based on some priority measures. Besides this, we can help the target user make the right decisions by suggesting the most likely values which can replace the obsolete ones with some prediction precision. This may be of great interest in further research.

\subsection{Analysis and validation of the robustness of our approach}

The comparison of experimental methods is essential to assess systematic errors and to study the effectiveness and robustness of the proposed approach in relation to existing work.
As mentioned in Sect.~\ref{section3}, the proposed approach is novel and there is little work close to ours. Currently, there is no readily available method to fairly compare it with our work. In Bayesian belief update-based methods \cite{gyenis:modal,schwering:belief}, authors update the knowledge base, which is not the object of our study. In \cite{mgns:tracing,livshits:shapley}, authors use propositional formulas to represent their knowledge. We therefore cannot apply this logic model since we work within an uncertain environment that requires uncertainty measures.
In \cite{chennovel}, authors propose a classification model to detect abnormal pedestrian trajectories. 
However, in our work, the investigation does not end with the detection of anomalies; rather, it includes identifying the input information responsible for these anomalies in order to update it.

A conceivable technique to demonstrate the effectiveness of our strategy is as follows: (1) we attempt to apply our OIDA on the same database of scenarios $\mathcal{S}_{Elderly}$, but with a BN that we learned from training data, (2) we try to apply our approach on two other real-life CBNs in two different contexts: diagnosis of liver disorder and monitoring patients in intensive care units.

\subsubsection{Results using some usual classifiers}

First, from the database \emph{Elderly-Data} and using a structure learning algorithm, we get a $41$-variable Bayesian model which is a DAG but not causal, denoted by $\mathcal{BN}_{Elderly}$. The CPTs are learned from empirical data in \emph{Elderly-Data} once the $\mathcal{BN}_{Elderly}$ is built. Since the detection of contradictions is based on a threshold, and since the threshold depends on the Bayesian model, we used the method described in Sect.~\ref{section6} to calculate the appropriate threshold relative to the learned $\mathcal{BN}_{Elderly}$. The threshold is set at $0.1$.

To better understand the effects of causality, we examine the FP rate and the FN rate of our method (using the $\mathcal{CBN}_{Elderly}$) in comparison to those obtained using the learned $\mathcal{BN}_{Elderly}$.
As can seen from Table~\ref{tabComp}, the accuracy of contradictory and non contradictory scenarios detection (step 2 in Fig.~\ref{fig1:agent}) of the proposed method ($93\%$) is better than those using a BN learned from \emph{Elderly-Data} ($48\%$).

\begin{table}[h]
    \centering
    \begin{tabular}{|M{2.8cm}|M{2.5cm}|M{2.5cm}|M{2cm}|}
         \hline
         Methods & Accuracy of contradictory scenarios detection & Accuracy of non-contradictory scenarios detection & Accuracy \\
         \hline
         using $\mathcal{BN}_{Elderly}$ & $35\%$ & $62\%$ & $48\%$ \\
         \hline
          using $\mathcal{CBN}_{Elderly}$ & $95.4\%$ & $92.3\%$ & $93.8\%$\\
         \hline
    \end{tabular}
    \caption{Results of OIDS using $\mathcal{CBN}_{Elderly}$ and BN}
    \label{tabComp}
\end{table}

Moreover, we also thought of applying other classifiers such as, decision tree, support vector machine, etc. in order to study their efficiency in detecting contradictory scenarios in comparison with our approach. However, given the nature of our problem and the type of information we manage, it was found that the other classifiers are inappropriate for many reasons. Indeed, among the 41 variables selected for this study, an arbitrary number of them can be observed for each scenario $S_i$, i.e., scenarios contained in the database $\mathcal{S}_{Elderly}$ do not have the same size (number of observed variables). This situation makes it very difficult to use usual classifiers because a new model would have to be learned for each possible subset of observed variables related to a specific subject, and it makes it almost impossible, because in real-life, we do not have enough scenarios composed of the same observed variables related to different subjects in order to train the mode. Moreover, 
the information about patients collected can be used by data-driven methods to inform clinical decision making. However, building machine learning methods to extract actionable intelligence (such as knowing exactly which observations are obsolete compared to others and updating them) from observational patient data involves causal model, which goes beyond standard supervised learning methods for prediction.

Causal Bayesian models allow us to overcome these problems since the  same model can  be used to evaluate any variable in the model, regarding any subset of observations. In addition, CBNs allow to combine general causal knowledge and specific individual information to analyze and interpret deep, valuable, and personalized situations such that the reasons and circumstances underlying unexpected outcomes even from incomplete observations.

\subsubsection{Results using two others CBNs}
\begin{sloppypar}
Another way to validate our approach is to apply our OIDA using two CBNs in two different application domains: the \emph{HeparII} model, denoted by $\mathcal{CBN}_{HeparII}$ for diagnosis of liver disorders; and the \emph{Alarm} model, denoted by $\mathcal{CBN}_{Alarm}$ for monitoring patients in intensive care units.
The experiments are carried out using two databases, $\mathcal{S}_{HeparII}$ and $\mathcal{S}_{Alarm}$, of contradictory and non contradictory scenarios related respectively to $\mathcal{CBN}_{HeparII}$ and $\mathcal{CBN}_{Alarm}$.
 
The $CBN_{HeparII}$ model is built as part of the project \emph{HEPAR} at the Institute of Biocybernetics and Biomedical Engineering of the Polish Academy of Science, in collaboration with physicians at the Medical Center of Postgraduate Education in Warsaw \cite{bobrowskihepar, oniskoextension}.
The structure of the $CBN_{HeparII}$ consists of $73$ nodes that represent the patient characteristics required for the diagnosis of livers disorder and is elicited from expert diagnosticians, while the parameters are learned from a database of medical cases. The CBN related to the \emph{HEPAR} project is available at: \url{http://aragorn.pb.bialystok.pl/~aonisko/}.
\end{sloppypar}

The $\mathcal{CBN}_{Alarm}$ model was developed to simulate causal relations in the emergency medical system \emph{Alarm} (A Logical Alarm Reduction Mechanism) \cite{beinlichalarm}. It  connects $8$ diagnosis, $16$ findings, and $13$ intermediate variables used for monitoring patients in intensive care units. The CBN associated with Alarm is available at: \url{https://www.bnlearn.com/bnrepository/discrete-medium.html\#alarm}.

Two sets of simulated data, $\mathcal{S}_{HeparII}$ and $\mathcal{S}_{Alarm}$ and two thresholds $\epsilon_{HeparII}$ and $\epsilon_{Alarm}$ were generated respectively from $\mathcal{CBN}_{HeparII}$ and $\mathcal{CBN}_{Alarm}$ using the protocol described in Sect.~\ref{section6}. 
The dataset $\mathcal{S}_{HeparII}$ (resp. $\mathcal{S}_{Alarm}$) contains $460$ (resp. $520$) scenarios divided equally into contradictory and non-contradictory scenarios labeled by an infectious diseases specialist and a radiologist (resp. an intensive care anesthetist). The threshold $\epsilon_{HeparII}$ (resp. $\epsilon_{Alarm}$) is set at $0.05$ (resp. $0.02$).

We apply our OIDA on the two datasets $\mathcal{S}_{HeparII}$ and $\mathcal{S}_{Alarm}$. The objective is to compare how our system performs compared to human diagnosticians (i.e. physicians) and whether its impact on detecting personal contradictory situations and identifying obsolete observations is beneficial. The results of the two experiments as well as those using our $\mathcal{CBN}_{Elderly}$ are summarized in Table~\ref{tab5}. As our system is composed of two parts: the detection of contradictions and the identification of obsolete observations, lines 7 and 8 of Table~\ref{tab5} represent, respectively, the accuracy of our system in each phase.

\begin{table}[h]
    \centering
    \begin{tabular}{|M{3.5cm}|M{2.2cm}|M{2.2cm}|M{2.2cm}|}
         \hline
          & $\mathcal{CBN}_{HeparII}$ & $\mathcal{CBN}_{Alarm}$ & $\mathcal{CBN}_{Elderly}$  \\
         \hline
          \textbf{Nb of variables} & $73$ & $37$ & $41$   \\
         \hline
          \textbf{$\epsilon$ threshold} & $0.05$ & $0.02$ & $0.01$  \\
         \hline
          \textbf{Nb of scenarios in the database $\mathcal{S}$} & $460$ & $520$ & $560$ \\
         \hline
          \textbf{Sensitivity} & $93.1\%$  & $89.5\%$ & $95.4\%$ \\
          \hline
          \textbf{Specificity} &  $94.4\%$  & $90.1\%$ & $92.3\%$ \\
          \hline
          \textbf{Accuracy} & $93.7\%$ & $89.8\%$ &  $93.87\%$ \\
          \hline
          \textbf{Accuracy of AND-OR Trees identification} &  $93.9\%$   & $89.5\%$ & $93.5\%$ \\
        \hline
    \end{tabular}
    \caption{Results of simulations on different CBNs.}
    \label{tab5}
\end{table}

The experiments on the three CBNs showed quite high detection and identification accuracy of obsolescence. 
The reaction of the physicians to the first trial version of our system was very favorable, and several of them said that working with the system is beneficial for them. Moreover, our system did not have any negative impact on the users, none of the good decisions made by the doctors were changed, even if the system provided an incorrect answer. It has led us to the conclusion that our system could be useful in assisting physicians and other users in detecting abnormal situations, understanding and identifying all possible causes, and deciding how to deal with these situations to prevent any kind of risk as early as possible.

Note that the error rate of our system comes from imperfections inherent in the CBN. Thus, some conclusions will be incorrect, no matter how carefully drawn. Furthermore, a better representation cannot save us: all representations are imperfect, and any imperfection can be a source of error.
Therefore, we always assume that all results are given with a degree of uncertainty.

\section{Conclusion}
\label{section8}
In this work, we proposed an entirely novel information-updating approach when a CBN is used as a representation model. The core idea is to identify obsolete information when it contradicts other newly acquired one in order to restore a database consistency. Our approach is theoretically supported, first, by the proposition of a new concept $\epsilon$-Contradiction to detect contradictions between a set of observations. Second, by defining and proving new properties and propositions to identify obsolete information efficiently. We design a polynomial-time algorithm to solve the information obsolescence problem. One of the most significant contributions is that the results are presented in an original way, in the form of an explanation AND-OR tree. It encodes all possible obsolete observations and can be effectively used to update information. We demonstrate the applicability of the proposed approach on a real database in the context of an European elderly fall-prevention project.
Our approach efficiency is confirmed experimentally since our simulations are very encouraging, reaching an accuracy of $93\%$.

One of the work's limitations is that the newly acquired information was assumed to be certain during processing. However, this may not be the case in real life, and this information may, therefore, make the recommendations inaccurate. We believe that our approach can be improved in future work by extending the update process to cover checking newly acquired information for fear of being the primary source of inconsistency. Furthermore, one possible direction to improve the results is by exploring the information aging by minutely studying the temporal correlation of each CBN variable.
We will also answer the following questions: how the resulting trees can be used to give reliable recommendations and how to remove obsolete information from the database to restore its consistency.
Finally, we aim to conceive a user interface in order to perform a set of tests of our OIDS by some physicians, using an iterative and incremental development cycle.
We consider the development of these ideas to be a promising avenue for future research.

\section*{Acknowledgements}
The present work is part of the ELSAT2020\footnote{http://www.elsat2020.org/en} project, which is co-financed by the European Union with the European Regional Development Fund, the French state and the Hauts de France Region Council. It is also supported by the PEJC project (20PEJC 08-03) fund from the Tunisian ministry of higher education
and scientific research. The experts who provided the estimates for the used causal Bayesian model and the University Hospital physicians who validated our scenarios are thanked for their participation.

\bibliographystyle{spmpsci}
\bibliography{bibliofile}


\end{document}